\documentclass[10pt]{article} 
\usepackage[preprint]{tmlr}

\usepackage{amsmath,amsfonts,bm}

\def\eqref#1{equation~\ref{#1}}

\def\1{\bm{1}}

\DeclareMathAlphabet{\mathsfit}{\encodingdefault}{\sfdefault}{m}{sl}
\SetMathAlphabet{\mathsfit}{bold}{\encodingdefault}{\sfdefault}{bx}{n}

\usepackage[utf8]{inputenc} 
\usepackage[T1]{fontenc}    
\usepackage{hyperref}       
\usepackage{url}            
\usepackage{booktabs}       
\usepackage{amsfonts}       
\usepackage{nicefrac}       
\usepackage{microtype}      
\usepackage{xcolor}         

\usepackage{graphicx} 
\usepackage{amsmath}
\usepackage{subcaption}

\title{Is ChatGPT Transforming Academics' Writing Style?}

\author{\name Mingmeng Geng \email mgeng@sissa.it \\
      \addr International School for Advanced Studies (SISSA), Trieste, Italy.
      \AND
      \name Roberto Trotta \email rtrotta@sissa.it \\
      \addr International School for Advanced Studies (SISSA), Trieste, Italy. \\
       Imperial College London, London, UK.}

\begin{document}

\maketitle
\begin{abstract}
Based on one million arXiv papers submitted from May 2018 to January 2024, we assess the textual density of ChatGPT's writing style in their abstracts through a statistical analysis of word frequency changes. Our model is calibrated and validated on a mixture of real abstracts and ChatGPT-modified abstracts (simulated data) after a careful noise analysis. The words used for estimation are not fixed but adaptive, including those with decreasing frequency. We find that large language models (LLMs), represented by ChatGPT, are having an increasing impact on arXiv abstracts, especially in the field of computer science, where the fraction of LLM-style abstracts is estimated to be approximately 35\%, if we take the responses of GPT-3.5 to one simple prompt, ``revise the following sentences'', as a baseline. We conclude with an analysis of both positive and negative aspects of the penetration of LLMs into academics' writing style.
\end{abstract}

\section{Introduction}
Since ChatGPT (Chat Generative Pre-trained Transformer) was released on November 30, 2022, large language models (LLMs) have started to be widely exploited and therefore affect many aspects of our lives. In this paper, we are concerned with whether LLMs, represented by ChatGPT, are transforming academic writing.

Many papers have already explored the advantages and disadvantages of LLMs~\citep{kasneci2023chatgpt}.
Although they can increase productivity and may help scientific discovery~\citep{noy2023experimental,ai4science2023impact}, the potential risks of using LLMs in academia cannot be ignored~\citep{lund2023chatgpt}, such as generating incorrect references~\citep{walters2023fabrication}.

Researchers have been working on machine-generated text detection since a few years ago~\citep{bakhtin2019real,gehrmann2019gltr}, which have attracted more attention shortly after ChatGPT appeared~\citep{mitchell2023detectgpt,guo2023close}. At the same time, questions have been raised about the reliability of these detectors~\citep{sadasivan2023can}. Detection and counter-detection of LLM-generated text soon developed cat-and-mouse games, such as watermarking~\citep{kirchenbauer2023watermark}, paraphrasing~\citep{sadasivan2023can}, and the combination of both~\citep{krishna2024paraphrasing}. Besides, distinguishing between human and LLM writing samples is also sometimes not easy for human experts~\citep{casal2023can}.

While there is already a corpus of current research on using ChatGPT in academia~\citep{casal2023can,lingard2023will,fergus2023evaluating,lund2023chatgpt}, to our knowledge only a handful of works have attempted to quantify its impact on the whole academic community. It was only when the first version of this paper was completed that two preprints appeared that addressed related questions: one focuses on AI conferences peer reviews~\citep{liang2024monitoring} and analyzes scientific papers~\citep{liang2024mapping}. They claim that the usage of LLMs is evident in AI conference reviews and scientific writings, especially in computer science papers. 

Within the broad field of academic writing and publishing, we chose the abstracts of articles as the focus of this work, as they have a relatively uniform format across disciplines, are supposed to condense an entire research article and thus are often highly polished, and can be considered short articles of pure text, not involving pictures nor tables.

Of course, LLMs can generate abstracts directly given a suitable prompt~\citep{luo2023chatgpt}, and studies have shown that identifying such abstracts is not easy even if they remain unedited by humans ~\citep{gao2023comparing,cheng2023comparisons}. We have previously discussed methods for detecting LLM-generated text, but the detection of a mixture of human and machine-generated text is usually much harder~\citep{krishna2024paraphrasing,zhang2024llm}. Determining whether a given few sentences were generated by LLMs is difficult, but it is feasible to estimate the extent to which millions of sentences are influenced by LLMs. We analyzed the fingerprints of LLMs on scientific abstracts as a function of time in order to tease out a statistical signature, rather than a binary classification.

In fact, that the abstract of a paper shows what we call the ``ChatGPT style'' or ``LLM style'' does not necessarily mean that the authors directly utilized LLM to generate or modify it. It is also possible that the authors used LLM in another context and that, as a result, their writing habits were influenced by the LLM style -- not a remote possibility. 

It is worth considering in this context that reading and writing in English is more difficult for non-native English academics~\citep{amano2023manifold}. Before ChatGPT was released, the pros and cons of other tools were discussed, such as Google Translate~\citep{mundt2016double} and Grammarly~\citep{fitria2021grammarly}, but ChatGPT has a much wider range of application scenarios -- not to mention, a much higher flexibility. 

We have seen similar AI-induced seismic shifts in the past: after AlphaGo~\citep{silver2017mastering} shocked the world, professional Go players have begun training with AI, and the sport of Go has been profoundly changed as a result~\citep{kang2022ai}. A similar story may be happening with academic writing, especially for researchers whose first language is not English~\citep{hwang2023chatgpt}. This paper is a first effort at establishing whether this is the case. 

We also think that analytical rigor is a higher priority than comprehensiveness, and the former is our focus in this paper. Once the reliability of a single analysis is assured, the comprehensive analysis can be more convincing. For example, we should use a more adaptive approach to selecting words for estimation, as well as considering words with decreasing frequency.

\section{Data}
\paragraph{arXiv dataset} The metadata of arXiv papers are provided by Kaggle~\citep{arXivDataset}. Because the abstracts in this dataset are updated when authors submit changes, we used the first version in 2024 (version 161) as well as the last version before the ChatGPT era (version 105). Our observations and analysis are based on one million arXiv articles submitted from May 2018 to January 2024. 

\paragraph{English word frequency} Google Ngram dataset is chosen for comparison and reference ~\citep{michel2011quantitative}. Specifically, we used the freely available mirrors on Kaggle (\url{http://kaggle.com/datasets/wheelercode/english-word-frequency-list}) covering word frequencies from the 1800s to 2019 as established from Google Books.

\section{Observations and analysis}
\subsection{Changes in word frequency}
\begin{figure}[h]
    \centering
    \begin{subfigure}[b]{0.48\textwidth}
        \centering
        \includegraphics[width=\textwidth]{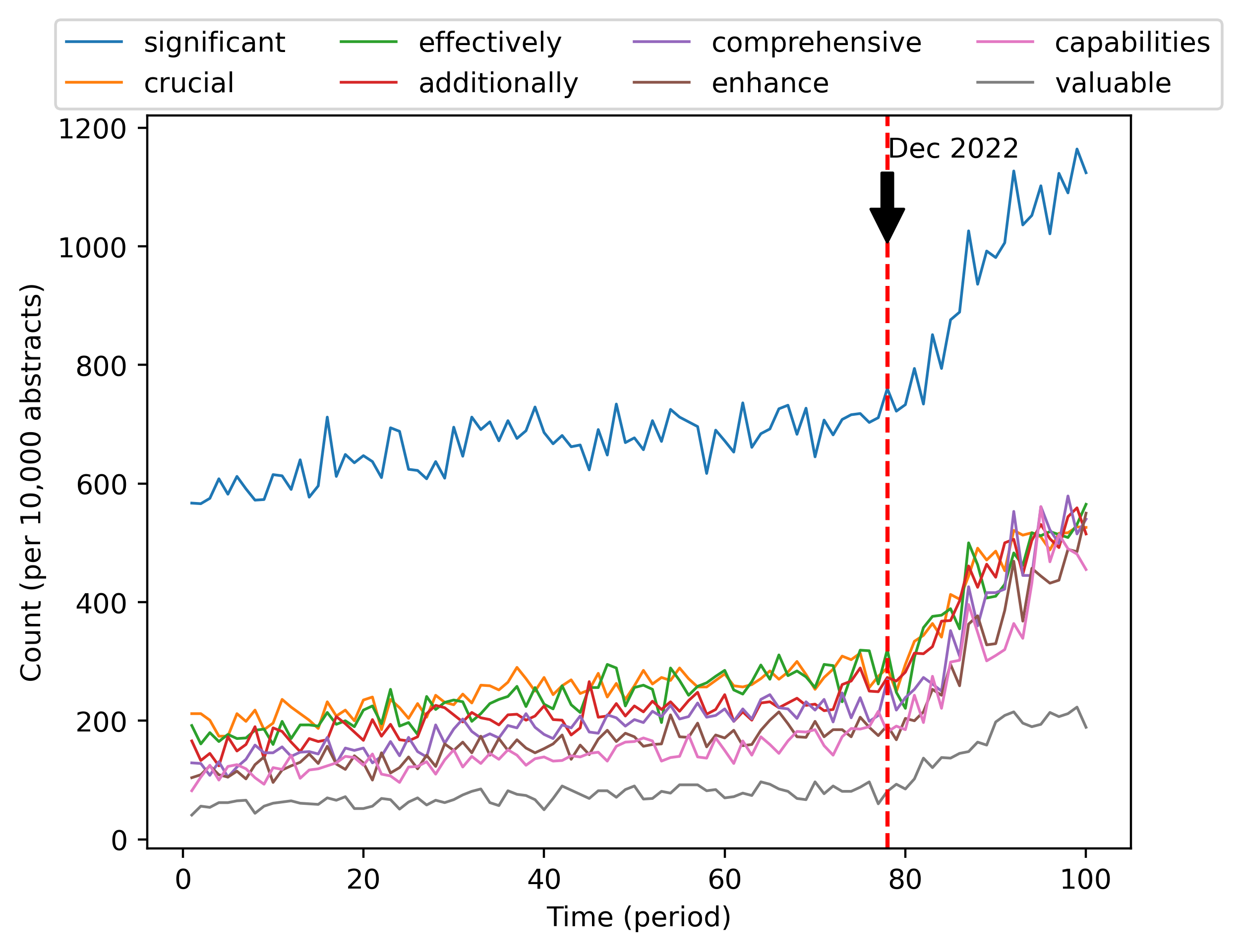}
        \caption{Examples of words with rapidly growing frequency in arXiv abstracts.}
        \label{ex_llms}
    \end{subfigure}
    \hfill
    \begin{subfigure}[b]{0.48\textwidth}
        \centering
        \includegraphics[width=\textwidth]{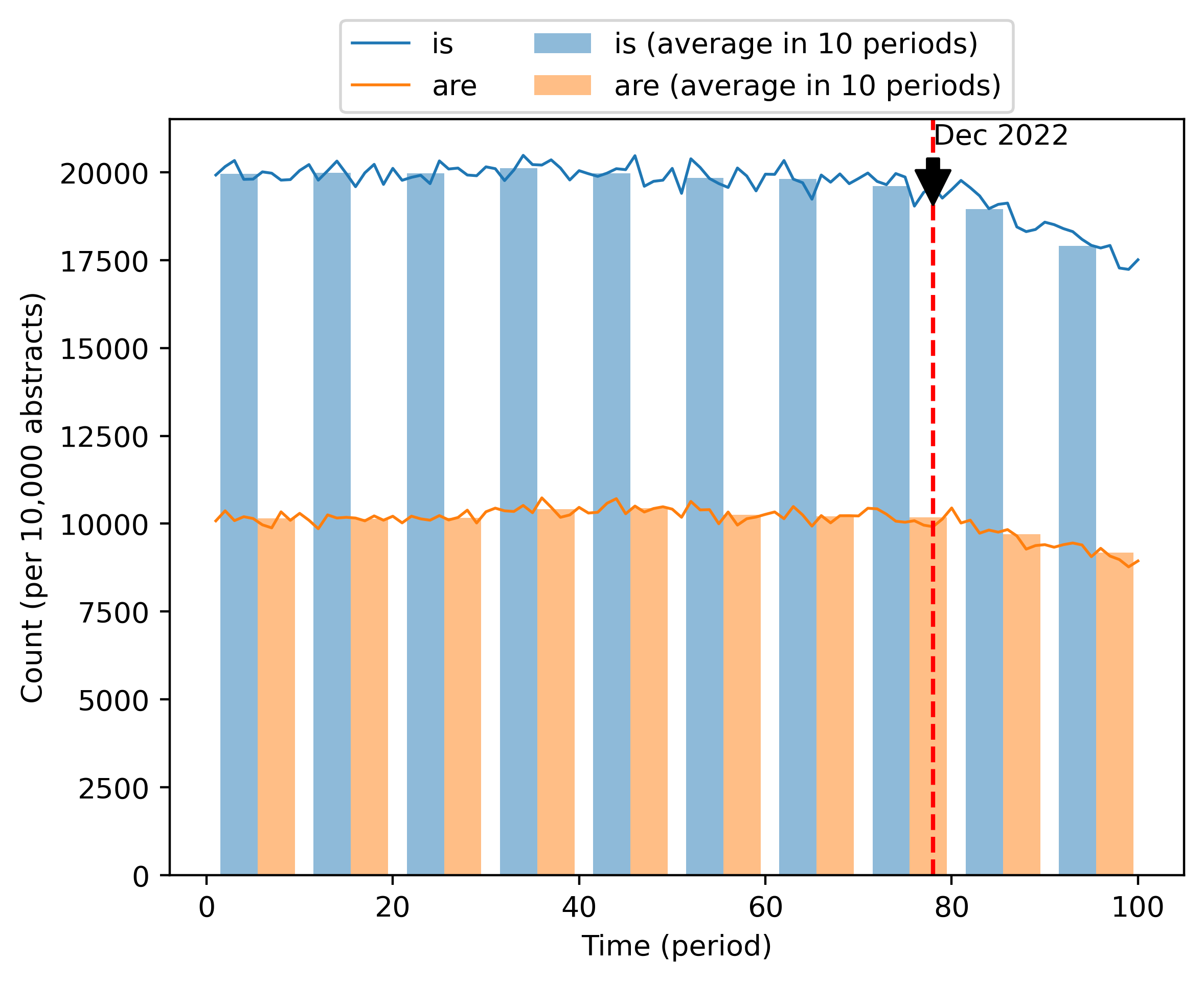}
        \caption{The words ``are'' and ``is'' are decreasing in frequency in arXiv abstracts.}
        \label{are_is}
    \end{subfigure}
    \caption{Word frequency changes in abstracts. The vertical red dashed line demarcates the first time period after ChatGPT's release.}
    \label{wc_change}
\end{figure}

We approach the problem by analyzing how the frequency of words changes after ChatGPT has been deployed. The frequency of some non-specialized words also starts to skyrocket in early 2023, as presented in Figure \ref{wc_change} (1 million abstracts are divided into 100 uneven time-periods, each encompassing 10,000 abstracts). The number of arXiv articles is getting more and more each month. In general, the larger the sample (the greater the number of articles), the more accurate the estimate will be. We used a similar number of articles in each period rather than the same time interval to keep the error in the estimates on the same scale, providing the same quality of observation and estimation.

How could the frequencies of words like ``\textbf{significant}'' grow \textbf{significantly} together?  Another striking example is the frequency change of the words ``are" and ``is''. The counts in 10,000 abstracts of these two words were quite stable before 2023. However, the frequency of these two terms has dropped by more than 10\% in 2023. 

These examples, anecdotal as they are, may represent the tip of the iceberg of a wider and growing phenomenon: the rapid increase in the usage of ChatGPT or other LLMs. The rise and fall in frequency of specific technical nouns may well be related to the changing popularity of certain research topics, but that a research trend is responsible for the change in usage of adjectives appears implausible -- even less so for words like ``is'' and ``are''.

\subsection{LLM simulations}

We wanted to be more specific about the impact of LLMs on articles from different disciplines, so the arXiv abstracts from different categories were examined separately. The one million arXiv articles were divided into 20 periods in this part in order to increase the number of articles per period and reduce estimation error, which is not the same as the previous part. The identifier numbers of the first and last arXiv articles corresponding to each period are given in the section \ref{periods} of the Appendix.

The emergence of other LLMs is also inspired or influenced by ChatGPT, and we also assume that other LLMs have similar but not identical word preferences to ChatGPT. The most recent articles we processed were submitted in January 2024, when other LLMs should be less utilized. Therefore, we used ChatGPT for simulations.

Previous studies have shown that ChatGPT has its own linguistic style~\citep{alafnan2023artificial}, and that likely includes the frequency of some words. Although there is no direct way to investigate ChatGPT's word preference, we can ask ChatGPT to polish or rewrite real, pre-2023 abstracts, and use the resulting simulation data to calculate the estimated frequency change rate $\hat{r}_{ij}$ of word $i$ in category $j$:
\begin{equation}
    \hat{r}_{ij} = \frac{\tilde{q}^d_{ij}-q^d_{ij}}{q^d_{ij}} =  \frac{\tilde{q}^d_{ij}}{q^d_{ij}} -1 \label{hat_r_ij}
\end{equation}
where $q^d_{ij}$ represents the word frequency of real abstracts in the dataset and $\tilde{q}^d_{ij}$ means the frequency after ChatGPT processing. 

What are the prompts used in the real cases are still unclear, and we think simple prompts could better reflect the inherent word preferences of ChatGPT, as complex prompts may bring more human interference. So some simple prompts were used to reduce the bias due to prompts, for example,
\begin{center}
\textit{``Revise the following sentences:''}
\end{center}

GPT-3.5 was utilized in our simulations for 10,000 abstracts in period 14 (April 2022 to July 2022), although it may not have the same word preferences as other LLMs. Many words have different frequencies before and after ChatGPT processing, such as the words ``is'', ``are'', and ``significant'' that we mentioned earlier. For simplicity, the results of the 4 categories with the highest number of articles are shown in Table \ref{wf_change} and the rest parts in this paper, namely \textit{cs} (computer science), \textit{math} (mathematics), \textit{astro} (astrophysics), and \textit{cond-mat} (condensed matter). 
\begin{table}[htb]
\caption{Word frequency (per abstract) before and after ChatGPT processing.}
\label{wf_change}
\begin{center}
\begin{tabular}{lccccr}
\toprule
words & category & before & after & change rate \\
\midrule
 is, are & cs  & 2.01, 1.00 & 1.73, 0.83 & -14\%, -17\% \\
 is, are & math &  1.78, 0.74 & 1.61, 0.71 & -9\%, -5\%  \\
 is, are & astro &  2.13, 1.39 & 1.90, 1.25 & -11\%, -1\% \\
 is, are & cond-mat & 2.00, 0.92  & 1.68, 0.80 & -16\%, -13\% \\
 significant & cs  & 0.09 & 0.18 & 99\% \\
 significant & math & 0.01  & 0.03  & 308\% \\
 significant & astro & 0.17  & 0.26 & 53\% \\
 significant & cond-mat & 0.07 & 0.18 & 171\%  \\
\bottomrule
\end{tabular}
\end{center}
\end{table}

This corroborates the hypothesis, formulated earlier, that the drop in the frequency of these two words observed in real abstracts in 2023 may have been caused by ChatGPT. Combined with Figure \ref{wc_change_com} in the Appendix showing the correlation between changes in simulated and real data, we speculate that ChatGPT is one of the important reasons, possibly even the main reason, for the recent word frequency change in abstracts. 

Our next step is to start by modeling LLM impact or ChatGPT impact, as well as estimating the impact based on real data and simulations. In order to minimize the impact of the research topic, different words should be used for estimation for different paper categories. And it is important to consider not only words that increase in frequency, but also those that decrease in frequency.

\section{LLM impact}
\subsection{Simple model}

Imagine different scenarios of using LLMs in scientific writing: a researcher might simply use it to correct grammatical errors, another employs it for translating native sentences into English, and yet another one wants it to polish their draft in English very purposefully. In theory, each of these use cases contributes the same proportion of LLM usage. But, as is well known, different prompts will lead to different outputs, which means different word frequency changes.
Therefore, we use the more neutral term ``LLM impact'' instead of ``proportion'' in our estimation part. Because the estimates in this paper are based on ChatGPT simulations, it can also be called ``ChatGPT impact''.

We start with a simple model, ignoring noise and variability for this subsection. Suppose that the frequency of word $i$ for abstracts in subject category $j$ changes from $f^*_{ij}$ to $\tilde{f}^*_{ij}$ after being processed by ChatGPT, when it's used as a means to polish and improve the abstract (if not to fully generate it). The corresponding word change rate is defined as
\begin{equation}
    \bar{r}_{ij}=\frac{\tilde{f}^*_{ij}-f^*_{ij}}{f^*_{ij}}=\frac{\tilde{f}^*_{ij}}{f^*_{ij}}-1 \, . \label{bar_r_ij}
\end{equation}
Suppose that $\bar{f}_{ij}(t)$ is the word frequency for word $i$ in category $j$ at time period $t$, this can be written as:
\begin{equation}
\bar{f}_{ij}(t) = (1-\eta_j(t)) f^*_{ij}(t) + \eta_j(t)f^*_{ij}(t)(\bar{r}_{ij}+1)= f^*_{ij}(t) + \eta_j(t)f^*_{ij}(t)\bar{r}_{ij}
\label{bar_f}
\end{equation}
where $\eta_j(t)$ denotes the proportion of abstracts in category $j$ affected by LLMs, and $f^*_{ij}(t)$ represents the original evolution in word frequency without LLMs. 

Unfortunately, we cannot know the true value of $f^*_{ij}(t)$ in the LLM era, but we can replace it with the estimation $\hat{f}^*_{ij}(t)$ based on the word frequency before LLM was introduced. As our objective is to identify the words that LLM ``likes'' (or ``dislikes'') to use compared to academic researchers on average, we assume that the frequencies of these words should remain stable without LLM, i.e., we take the average of the pre-ChatGPT periods before $T_0$ as following:
\begin{equation}
    f^*_{ij}(t) = \frac{1}{\#\{t\leq T_0\}} \sum_{t\leq T_0}f^d_{ij}(t),\text{if } t > T_0 \, . \label{fij_to}
\end{equation}

For a specified word $i$, we will have one estimate of $\eta_j(t)$, as $\bar{r}_{ij}$ and $f^*_{ij}(t)$ could be approximated with Eq.~(\ref{hat_r_ij}) and Eq.~(\ref{fij_to}). We are also likely to get better results after combining the estimates of different words.

However, this model is highly idealized: we have to additionally consider the effects of noise (such as randomness inside LLM), uncertainty in word usage evolution without LLM, and the epistemic uncertainty in how users actually prompt LLMs. 

\subsection{Noise model}
We now consider the noise terms, which might be modeled in many different ways. 

For instance, we denote the word frequency for word $i$ in category $j$ by ${f}^d_{ij}$, which represents the word frequency observed in the data:
\begin{equation}
    f^d_{ij} = f^*_{ij} + \delta_{ij}(f^*_{ij})
\end{equation}
where $\delta_{ij}(\cdot)$ represents noise and word usage variability which are not directly related to the internal parameters of LLM.

After taking into account the impact of LLMs, we split the word frequencies $\displaystyle f^d_{ij}(t)$ into two parts, $\displaystyle f^{\delta,\eta}_{ij}(t)$ and $\displaystyle f^{\delta,1-\eta}_{ij}(t)$, while they both have corresponding noise terms: 
\begin{align}
    f^{\delta,\eta}_{ij}(t) &= \eta_j(t)f^*_{ij}(t)+ \delta_{ij}(\eta_j(t)f^*_{ij}(t)) \\
    f^{\delta,1-\eta}_{ij}(t)&=(1-\eta_j(t)) f^*_{ij}(t) + \delta_{ij}((1-\eta_j(t)) f^*_{ij}(t)) \, .
\end{align}
In this case, the equation corresponding to Eq. (\ref{bar_f}) is
\begin{equation}
    f^d_{ij}(t) = (1-\eta_j(t)) f^*_{ij}(t) + \delta_{ij}((1-\eta_j(t)) f^*_{ij}(t)) + \mathrm{C}_{ij}(f^{\delta,\eta}_{ij}(t)) = f^{\delta,1-\eta}_{ij}(t) + \mathrm{C}_{ij}(f^{\delta,\eta}_{ij}(t)) \label{f_d_ij}
\end{equation}
where the function $\mathrm{C}_{ij}(\cdot)$ means the frequency after LLM process.

We assume that the noise in the ``real'' data and in the simulations due to LLM processing can be represented as $\epsilon_{ij}(\cdot)$ and $\epsilon_{ij}^s(\cdot)$, then Eq. (\ref{hat_r_ij}) and Eq. (\ref{bar_r_ij}) are related by
\begin{equation}
    \frac{\tilde{f}^*_{ij}-\epsilon_{ij}(f^*_{ij})-f^*_{ij}}{f^*_{ij}} = \frac{\tilde{q}^d_{ij}-\epsilon^s_{ij}(q^d_{ij})-q^d_{ij}}{q^d_{ij}} \, .
\end{equation}

Therefore,
\begin{equation}
    \mathrm{C}_{ij}(f^{\delta,\eta}_{ij}(t)) = f^{\delta,\eta}_{ij}(t)(\hat{r}_{ij} + 1 + \epsilon^{\eta}_{ij}(q,f,t))
\end{equation}
where
\begin{equation}
    \epsilon^{\eta}_{ij}(q,f,t) = \frac{\epsilon_{ij}(f^{\delta,\eta}_{ij}(t))}{f^{\delta,\eta}_{ij}(t)} - \frac{\epsilon^s_{ij}(q^d_{ij})}{q^d_{ij}} \, . \label{ep}
\end{equation}
Then, Eq. (\ref{f_d_ij}) -- representing the difference in word frequency before and after LLM processing -- can be rewritten as
\begin{equation}
        f^d_{ij}(t) - f^*_{ij}(t) =  \eta_{j}(t)x_{ij}(t) + g_{ij}(t) + \xi_{ij}(t) \label{f_ori}
\end{equation}
where
\begin{align}
x_{ij}(t) = &f^*_{ij} (t)\hat{r}_{ij} \label{x_ij}\\
g_{ij}(t) = & \eta_j(t)f^*_{ij}(t)\epsilon^{\eta}_{ij}(q,f,t) \label{g_ij} \\
    \xi_{ij}(t) =& (\hat{r}_{ij}+1+\epsilon^{\eta}_{ij}(q,f,t))\delta_{ij}(\eta_j(t)f^*_{ij}(t))+\delta'_{ij}((1-\eta_j(t)) f^*_{ij}(t))  \, . \label{xi}
\end{align}
where $\delta'_{ij}(\cdot)$ follows the same distribution as $\delta_{ij}(\cdot)$.
It should be noted that $g_{ij}(t)$ includes only LLM-related noise $\epsilon_{ij}(\cdot)$ and $\epsilon^s_{ij}(\cdot)$, however $\xi_{ij}(t)$ contains $\delta_{ij}(\cdot)$ and $\delta'_{ij}(\cdot)$ that are unrelated to LLM.

\subsection{Impact estimation and bias analysis}
In many data analysis applications, more data point (in our case, using a larger number of words) means better estimates. But in our case, the effect of noise is different for each data point (word), and choosing wisely which words to include can improve our estimates.

For simplicity, we define
\begin{equation}
    h_{ij}(t) = f^d_{ij}(t) - f^*_{ij}(t) \, .
\end{equation}
For abstracts in category $j$, we use the words in the subset $I_j$ (whose determination is discussed below), of numerosity $n_j$. In order to estimate $\eta_j(t)$, we can use the quadratic loss function
\begin{equation}
        L_{j,t}(\eta_j) = \frac{1}{n_j}\sum_{i \in I_j} ( h_{ij}(t) - \eta_{j}(t)x_{ij}(t))^2 =\frac{1}{n_j}\sum_{i \in I_j} (g_{ij}(t) + \xi_{ij}(t))^2 \, .
\end{equation}
If we ignored the dependency of $g_{ij}(t)$ and $\xi_{ij}(t)$ on $\eta_j(t)$, the estimate of LLM impact would simply be given by Ordinary Least Squares (OLS) as
\begin{equation}
    \hat{\eta}_j(t) = \frac{\sum_{i \in I_j} h_{ij}(t)x_{ij}(t)}{\sum_{i \in I_j} x^2_{ij}(t)} \, . \label{eta_hat_j}
\end{equation}
However, since $g_{ij}(t)$ also depends on $\eta_j(t)$ and $\xi_{ij}$ contains $\eta_j(t)$ as described in Eq. (\ref{g_ij}) and Eq. (\ref{xi}), we need to make additional assumptions to progress further.

\textbf{Case 1:} if the effect of $\eta_j(t)$ on $\xi_{ij}(t)$ can be ignored compared to other terms, e.g., the following simple scenario,
\begin{equation}
    \mathrm{Var}[\delta_{ij}(\eta_j(t)f^*_{ij}(t))] \ll \eta_j(t)f^*_{ij}(t) \mathrm{Var}[\epsilon^{\eta}_{ij}(q,f,t)]
\end{equation}
One can also derive the approximation below:
\begin{equation}
    f^{\delta,\eta}_{ij}(t) \approx \eta_j(t)f^*_{ij}(t)+ \delta_{ij}(*)
\end{equation}
where $\delta_{ij}(*)$ is a random variable with zero mean and variance much smaller than $\eta_j(t)f^*_{ij}(t)$, and its derivative with respect to $\eta_j(t)$ is negligible compared to $f^*_{ij}(t)$.

Therefore, the loss function under this assumption is:
\begin{equation}
        L_{j,t,g}(\eta_j) 
        = \frac{1}{n_j}\sum_{i \in I_j} ( h_{ij}(t) - \eta_{j}(t)x_{ij}(t)-g_{ij}(t))^2 \\
        = \frac{1}{n_j}\sum_{i \in I_j}\xi^2_{ij}(t) \, .
\end{equation}
Thus,
\begin{equation}
    \begin{split}
        \frac{\partial L_{j,t,g}(\eta_j)}{\partial \eta_j}=&\frac{2}{n_j}\sum_{i \in I_j}\left(\eta_j(t)x^2_{ij}(t)-h_{ij}(t)x_{ij}(t)\right) +\frac{2}{n_j}\sum_{i \in I_j} x_{ij}(t)g_{ij}(t) \\
        &-\frac{2}{n_j}\sum_{i \in I_j} \frac{\partial g_{ij}(t)}{\partial\eta_j(t)} \left( h_{ij}(t) - \eta_{j}(t)x_{ij}(t)-g_{ij}(t)\right) \,
    \end{split}
\end{equation}
If we require a minimum by setting $\frac{\partial L_{j,t,g}(\eta_j)}{\partial \eta_j} =0$, we obtain a new estimate $\hat{\eta}^g_j(t)$, which is equal to the OLS $\hat{\eta}_j(t)$ in Eq. (\ref{eta_hat_j}) corrected for bias and noise,
\begin{equation}
    \begin{split}
       (\hat{\eta}^g_j(t) - \hat{\eta}_j(t))\sum_{i \in I_j} x^2_{ij}(t) = &\sum_{i \in I_j} \frac{\partial g_{ij}(t)}{\partial\eta_j(t)} \left( h_{ij}(t) - \eta_{j}(t)x_{ij}(t)\right)  \\ 
        & - \sum_{i \in I_j} x_{ij}(t)g_{ij}(t) -\sum_{i \in I_j}g_{ij}(t)\frac{\partial g_{ij}(t)}{\partial\eta_j(t)} \, . \label{bias_eta_g}
    \end{split}
\end{equation}

But without knowing the distribution of $\epsilon_{ij}(\cdot)$ and $\epsilon^s_{ij}(\cdot)$, we have no way of estimating the value of this bias, so we assume that $\epsilon_{ij}(f_{ij}) \sim \mathcal {N}(0, f_{ij} \sigma_{ij,\epsilon}^{2})$ and $\epsilon^s_{ij}(f_{ij}) \sim \mathcal {N}(0,f_{ij} \sigma_{ij,\epsilon}^{2})$, e.g., $\epsilon_{ij}(1) \sim \mathcal {N}(0, \sigma_{ij,\epsilon}^{2})$, then we can obtain an expression for $\epsilon^{\eta}_{ij}(q,f,t)$: 
\begin{align}
\epsilon^{\eta}_{ij}(q,f,t) = & \frac{\epsilon_{ij}(1)}{\sqrt{\eta_j(t)f^*_{ij}(t)+ \delta_{ij}(*)}} - \frac{\epsilon^s_{ij}(1)}{\sqrt{q^d_{ij}}} \\
    g_{ij}(t) = & \frac{\eta_j(t)f^*_{ij}(t)\epsilon_{ij}(1)}{\sqrt{\eta_j(t)f^*_{ij}(t)+ \delta_{ij}(*)}}  - \frac{\eta_j(t)f^*_{ij}(t)\epsilon^s_{ij}(1)}{\sqrt{q^d_{ij}}} \, .
\end{align}
After calculations (see appendix), the bias part is expressed as
\begin{equation} 
\hat{\eta}_j(t) - \hat{\eta}^g_j(t) = 
    \frac{\sum_{i \in I_j}\mathrm{E}\left[g_{ij}(t) \frac{\partial g_{ij}(t)}{\partial\eta_j(t)}\right]}{\sum_{i \in I_j} (f^*_{ij} (t)\hat{r}_{ij})^2} \, . \label{eta_bias_e}
\end{equation}

Some insights can be gained from the results above. As by definition $\eta_j(t)\geq 0$, the estimate $\hat{\eta}_j(t)$ given by Eq. (\ref{eta_hat_j}) tends to be biased high in our model. The value of $\hat{r}_{ij}$ plays a role in the minimization of bias, as it only appears in the denominator in Eq. (\ref{eta_bias_e}). 

Similarly, if the value of $\hat{r}_{ij}$ is similar for different words, then larger values of $q^d_{ij}$ and $f^*_{ij}$ will reduce the bias, as seen from Eq. (\ref{eq:eta_bias_e_expression}) -- therefore, we should consider including preferentially in our analysis words with larger values of $q^d_{ij}$, $f^*_{ij}$ and $|\hat{r}_{ij}|$. Considering that the value of $\eta_j(t)$ affects the bias as well, which is not simply linear, we are led to consider adaptive or iterative criteria for word choice, which will in general depend on the true (and unknown) value of $\eta_j(t)$.

\textbf{Case 2:} Gaussian distribution for $\delta_{ij}(f_{ij})$, e.g., $\delta_{ij}(f_{ij}) \sim \mathcal {N}(0 ,f_{ij} \sigma_{ij}^{2})$, which is inspired by central limit theorem and justified empirically in the Appendix, Figure \ref{var_mean}.
As a result,
\begin{equation}
    \begin{split}
    \xi_{ij}(t) =&(\hat{r}_{ij}+\epsilon^{\eta}_{ij}(q,f,t))\delta_{ij}(\eta_j(t)f^*_{ij}(t))  +\delta'_{ij}(f^*_{ij}(t))  \\
    =& \sqrt{\eta_j(t)f^*_{ij}(t)}(\hat{r}_{ij}+\epsilon^{\eta}_{ij}(q,f,t))\delta_{ij}(1) +\sqrt{f^*_{ij}(t)}\delta'_{ij}(1) 
    \end{split}
\end{equation}
which gives us similar conclusions. (Some calculations are in the Appendix.)

Finding criteria for selecting the words that are included in the frequency change analysis greatly reduces the computational complexity compared to trying different word combinations. If all combinations of $n$ words are tried, that complexity rises to $O(2^n)$. When we use word choice criteria to select several groups of words, the complexity is reduced to $O(1)$. Our analysis of noise models gives some insights into these criteria, such as $q^d_{ij}$ and $\hat{r}_{ij}$.

\subsection{Calibration and test}
In order to verify the theoretical and practical validity of our approach, we used calibrations and tests, with ChatGPT-processed abstracts mixed with real abstracts. Considering that the noise in real data is likely highly complex, we did not estimate the variance of $\epsilon_{ij}(\cdot)$. Instead, we used ChatGPT to process additional abstracts (on top of those used to estimate $r_{ij}$), and used the resulting frequencies as calibration for the bias and noise. 

As previous analyses have demonstrated, with the goal of reducing bias in estimation, different selected words are likely to correspond to the different (unknown) ground truth value of $\eta_j(t)$. Therefore, we construct $N$ different sets of abstract data for calibration and test, $D_n$ and $T_{n'}$, with its correspond mixed ratio of ChatGPT-processed abstracts, $\eta_n$ and $\eta'_{n'}$, as
\begin{equation}
    (D_n,\eta_n), n\in \{1,2,\dots,N\}; \quad (T_n',\eta_n'), n'\in \{1,2,\dots,N'\}  .
\end{equation}

And for one pair of $(D_n,\eta_n)$ and a specific word choice requirement $q_k$ (for example, $q^d_{ij}>0.1$ and $\displaystyle \frac{\hat{r}_{ij}+1}{\hat{r}^2_{ij}} <\frac{0.1+1}{0.1^2}$), the efficiency can be defined as
\begin{equation}
    e(D_n, \eta_n,q_k) = |\eta_n-\hat{\eta}_n(D_n,q_k)|
\end{equation}
where $\hat{\eta}_n(D_n,q_k)$ is the estimate of $\eta_n$ using Eq. (\ref{eta_hat_j}) and the words set $I_j$ can be derived from $q_k$, denoted $I_j(q_k)$.

For a given set of $q_k$ (examples can be found in the Appendix), we are looking for the best one minimizing $e(D_n, \eta_n,q_k)$, denoted $q(D_n,\eta_n)$, which is the calibration part.  For the test data $T_{n'}$, the estimate of $\eta_{n'}$ is calculated from Eq. (\ref{eta_hat_j}) with different $I_j$, based on different $q(D_n,\eta_n)$ obtained in the calibration procedure. 

Because of the goal of the calibration, word choice may well actually introduce a new bias to neutralize the original bias, so that the estimate is not necessarily higher in the test results than the ground truth.

\section{Results}
\subsection{Calibration and test results}
To calibrate the choice of set $I_j$, we use different mixing ratios, in proportion to the value of $\eta_j(t)$. In addition, we only consider the 10,000 words with the highest frequency in the Google Ngram dataset.

We continue our simulations based on GPT-3.5. As the training data for GPT-3.5 is up to September 2021, abstracts submitted later than this time are considered: 20,000 abstracts in period 13 to estimate $r_{ij}$, 10,000 abstracts in period 12 for calibration, and 10,000 abstracts in period 14 for testing.

We used the first 10 periods before ChatGPT was introduced, to estimate $f^*_{ij}(t)$, as they weren't influenced by ChatGPT, which means $T_0=10$ and $\displaystyle \#\{t\leq T_0\}=10$ in Eq. (\ref{fij_to}).

We take $\displaystyle \{\eta_n\}=\{0,0.05,0.1,\dots,0.45,0.5\}$ and $m=1$, which means $N=\#\{(D_n,\eta_n)\}=11$.
Then the 11 $I_j$ (with possible repetitions), obtained from mixed data with 11 corresponding $\eta_n$ of period 12, were used for $\eta_n'$ estimation in the test data (period 14). Other parameters can be found in the Appendix. 

\begin{figure}[h]
    \centering
    \begin{subfigure}[b]{0.3\textwidth}
        \centering
        \includegraphics[width=\textwidth]{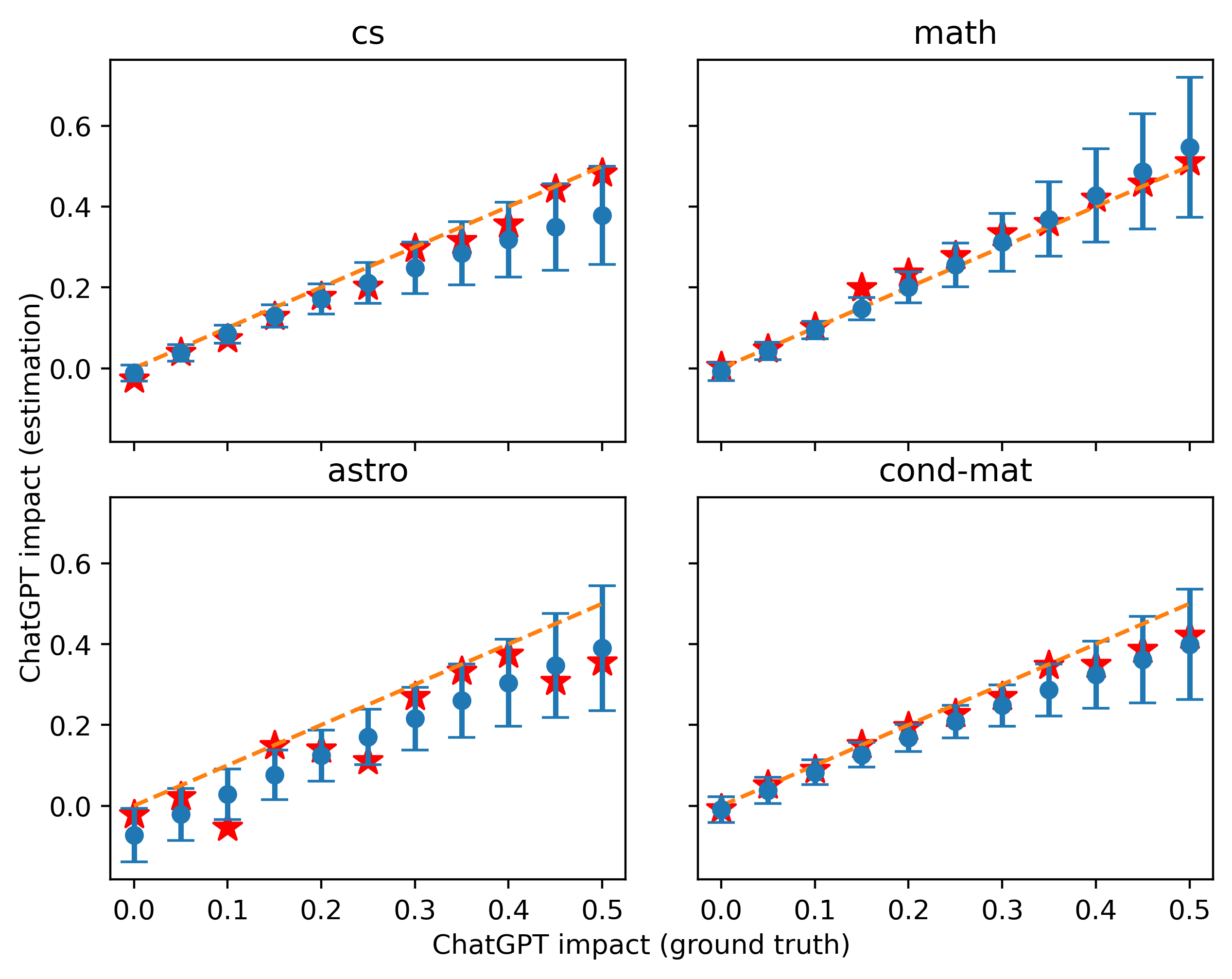}
        \caption{Normalized to the total number of abstracts.}
        \label{cal_test}
    \end{subfigure}
    \hfill
    \begin{subfigure}[b]{0.3\textwidth}
        \centering
        \includegraphics[width=\textwidth]{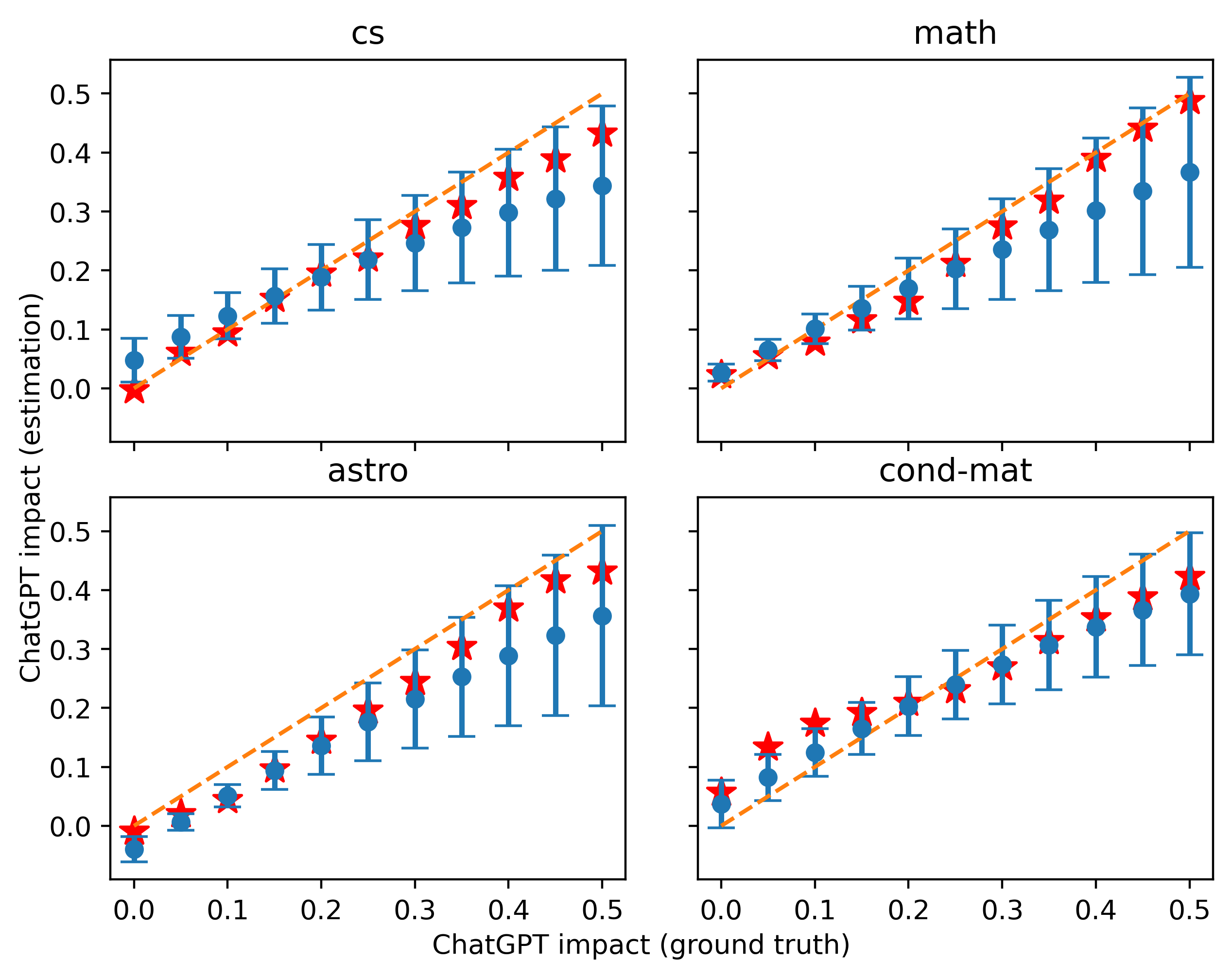}
        \caption{Normalized to the total number of words.}
        \label{cal_test_norm}
    \end{subfigure}
    \hfill
    \begin{subfigure}[b]{0.3\textwidth}
        \centering
        \includegraphics[width=\textwidth]{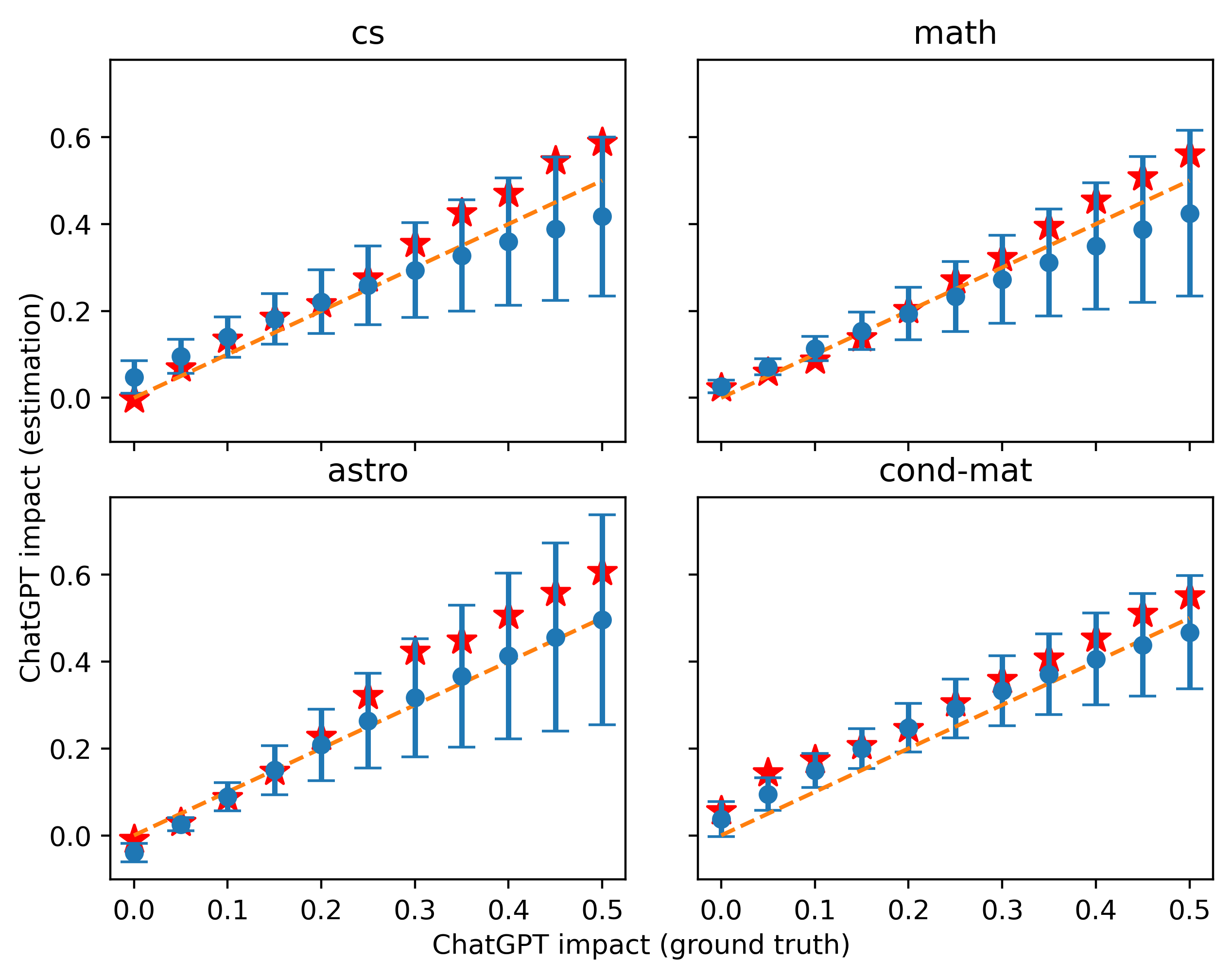}
        \caption{Different prompt for test data than used in calibration data.}
        \label{cal_test_norm_dif}
    \end{subfigure}
    \caption{Test results for simulated admixtures of abstracts in period 14. The error bars represent the standard deviation of the estimation results, and the red star is the estimated value of $\eta_n'$ from test data based on optimal $I_j$ with the same mixed ratio $\eta_n$ as in the calibration data. The orange dashed lines correspond to perfect estimation.}
    \label{cal_test_all}
\end{figure}

The results using the same prompt for generating calibration and test data are shown in Figure \ref{cal_test}, with injected mixed ratio (i.e., ChatGPT impact) $\eta_n'$ from 0 to 0.5. It is clear that when the calibration and test sets are mixed in the same ratio, word combinations that achieve better estimates on the calibration set generally work better on the test set, as well.  

Unlike in Figure \ref{cal_test} where we normalized the word frequency by the total number of abstracts, we normalized it by the total number of words for one period in Figure \ref{cal_test_norm}. The trends remain similar, albeit different in detail.

Because one may use a wide variety of prompts in practical applications, we also evaluated the robustness of our approach by adopting a different prompt for generating the test data than the one we used for calibration. The corresponding results in Figure \ref{cal_test_norm_dif} use the following prompt:
\begin{center}
\textit{"Please rewrite the following paragraph from an academic paper:"}
\end{center}
In this example, we add the word ``please'' and make it clear that this comes from an ``academic paper'', replacing ``revise'' with ``rewrite''. Although the quantitative results of our tests were not as good as before, the errors were still small at lower mixed ratios, which also illustrates the robustness of our method. This is understandable because in data generated with different prompts, not all of our previous assumptions hold, and the estimate of $\hat{r}_{ij}$ on $r_{ij}$ in our model may be biased. We can also note that most of our estimates in Figure \ref{cal_test_norm_dif} are on the high side relative to the ground truth, most likely because we use a more precise prompt for the test data here, making the frequency change rate of the relevant words higher.

\subsection{Estimation from real data}
\begin{figure}[h]
    \centering
    \begin{subfigure}[b]{0.48\textwidth}
        \centering
        \includegraphics[width=\textwidth]{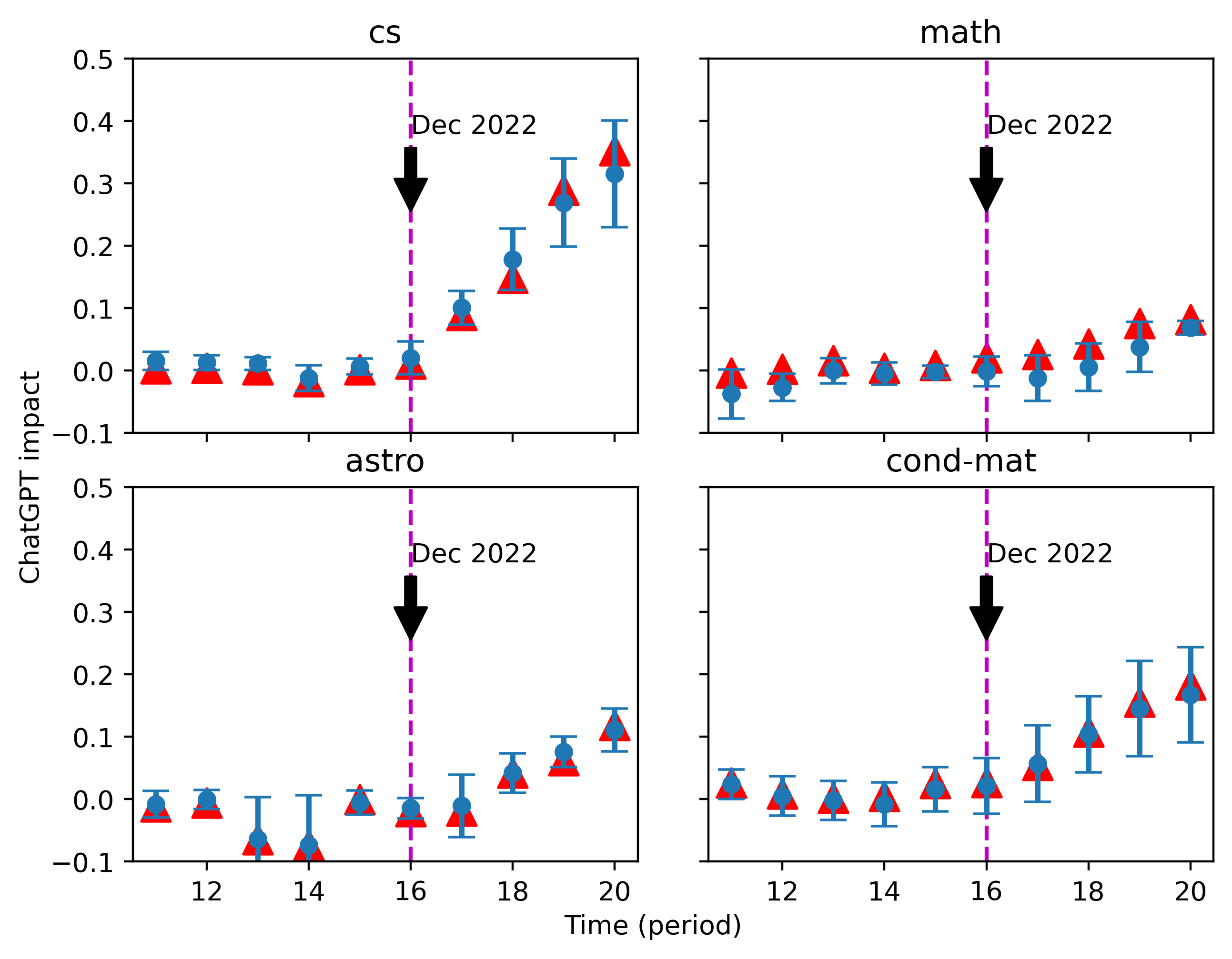}
        \caption{Normalized to the total number of abstracts.}
        \label{eta_10p}
    \end{subfigure}
    \hfill
    \begin{subfigure}[b]{0.48\textwidth}
        \centering
        \includegraphics[width=\textwidth]{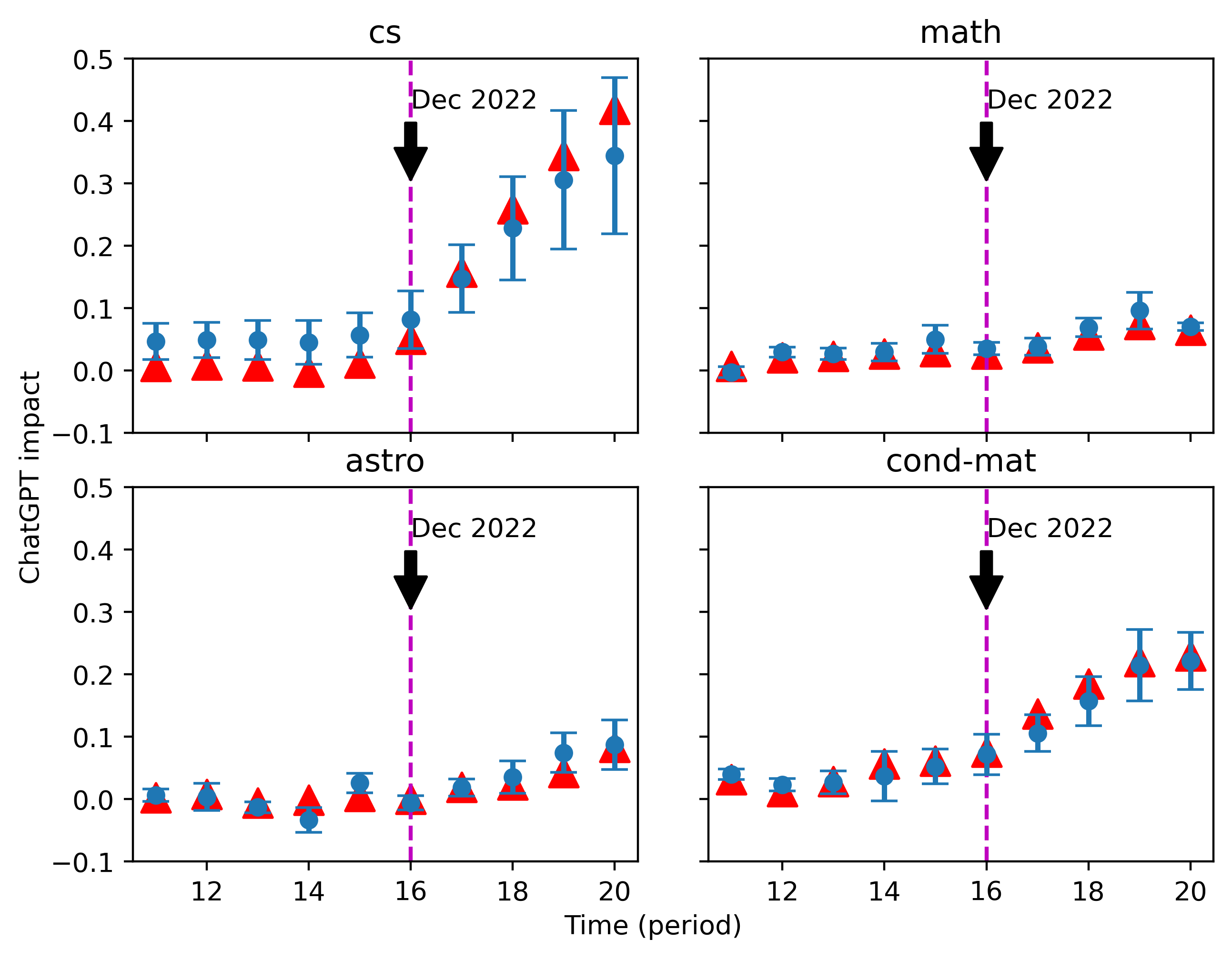}
        \caption{Normalized to the total number of words.}
        \label{eta_10p_norm}
    \end{subfigure}
    \caption{Estimates of $\eta_j(t)$ (i.e., ChatGPT impact) from real data. Word frequencies were normalized on the number of abstracts in each period before the estimation was performed. The error bars represent the standard deviation of the estimation results, using 11 different word sets $I_j$ obtained in the calibration procedure with 11 different $\eta_n$. The points of the triangle represent the average of the 3 estimates, corresponding to the 3 word selection requirements $q$ based on the 3 $\eta_n$ closest to the mean of the previous 11 estimates.}
    \label{eta_10p_all}
\end{figure}

The estimates of ChatGPT impact on the real data are shown in Figure \ref{eta_10p} and Figure \ref{eta_10p_norm}.  

Based on our calibration results, we chose 11 words set $I_j$ for different injected values of $\eta_n$. According to the results of the first estimation about $\eta_j(t)$, we found the three values of $\eta_n$ that were closest to the mean of the first estimation and used their optimal word set $I_j$ in the calibration procedure for a second estimation, leading to the triangle points shown in the figures. 

Despite mild differences in the estimates under the two different normalizations, the conclusions are essentially the same. Our estimates on $\eta_j(t)$ hover around 0 until 2023, which gives reassurance of the reliability of our methodology. More and more abstracts are being influenced by ChatGPT, especially in the \textit{cs} category, starting from December 2022, after the release of ChatGPT. 

Our estimate indicates that the density of ChatGPT style texts of the most recent time period in this category is around 35\%, when we use the results of one simple prompt, ``revise the following sentences'', as a baseline. By contrast, we detected a much smaller uptick in ChatGPT impact in \textit{math}, while \textit{astro} and \textit{cond-mat} both reach values between 10\% and 20\%, approximately.

It is important to note that our ChatGPT impact or LLM impact here is a relative value that corresponds to the change in word frequency from the use of simple prompts. More precise prompts, both in reality and in simulation, could potentially lead to an impact value greater than 1.  

\section{Conclusions}
Is ChatGPT transforming academics' writing style? An important question before these discussions is the evaluation of the actual penetration of the usage of ChatGPT in academic writing -- without a quantitative estimate, the debate is founded on anecdotal evidence. 

We have demonstrated here that we can monitor the impact of LLMs in arXiv abstracts by using simple and transparent statistical methods (e.g., word frequencies), which is easily extendable to other subjects and to the complete text of articles.

Some formulas above look complicated, but with the help of the calibration, the final estimates are linear regressions, i.e., Eq. (\ref{eta_hat_j}). And those formulas and proofs tell us which words should be theoretically selected for estimation, which to our knowledge other articles haven't done. In addition, we also propose adaptive word selection methods that are operationally simple.

Our estimates are founded on a population level and based on the output of simple prompts. Using more precise prompts, it is entirely possible to achieve abstracts that are more ChatGPT-like (or LLM-like) texts than our simulations. In addition, in the real world people might use LLMs other than ChatGPT to revise articles, which may have similar but not identical word preferences to ChatGPT.

We found convincing evidence of a change in word frequency after ChatGPT's release, consistent with predictions obtained from simulating LLM impact from possible users' prompts. The most enthusiastic community (among the four we investigated) in terms of LLM adoption appears to be that of computer scientists, a result that is perhaps unsurprising. Mathematicians, by contrast, are the least keen.

Our paper illustrates the importance of words chosen. Different types of articles with different LLM impact need to be estimated using the corresponding words, which we proved theoretically under certain assumptions and verified with simulated data. Not only did we focus on words that were increasing in frequency, but we also took words that were decreasing in frequency, which are not covered in other papers.

\section{Discussion}
The debate around the usage of generative models such as ChatGPT in academic writing is multi-faceted: from fears of lowering rigour due to ``hallucinations'' to uncertainty about the actual sources of AI-produced text. It is however indisputable that LLM tools such as ChatGPT also have positive impacts: they help non-English native writers to improve the quality and flow of their text, as well as to translate into English from their mother tongue or vice versa. In this sense, generative AI is a great leveller, and as such it is a welcome addition to the academic's toolbox. What we need to be wary of is its use in fully generative mode, without expert human supervision -- something that we have not addressed in this paper. 

We are aware that our methods can be further improved. For example, our results follow from analyzing a set of words selected based on the value of $q^d_{ij}$ and $\hat{r}_{ij}$. It is actually possible to fine-tune this criterium for a more accurate word selection, which would theoretically give better results, but would be more computationally expensive. Similarly, trying a larger range of prompts should theoretically result in better estimates. And better estimates may be made by more rigorous analysis, such as considering more complex noise terms. We are more interested in the density of LLM-style texts and its relative value (comparisons between categories and over time) than in establishing how many people are using LLMs -- this can be estimated with the help of questionnaires, and it is not possible to get an accurate estimate only based on simulated data.

As our results have shown, that LLMs, started by ChatGPT, are having an increasing impact on academic publications. This trend is hard to avoid, and we need to adapt gradually. With the increasing influx of young researchers, especially non-native English speakers, LLM tools represented by ChatGPT, are transforming academic writing, at least for some disciplines. Even if you refuse to use them, you are likely to be influenced indirectly.

\bibliography{paper}
\bibliographystyle{tmlr}

\newpage
\appendix

\section{Period divisions}
\label{periods}
\begin{table}[h]
\caption{First and last arXiv paper identifier of 20 periods.}
\label{per}
\begin{center}
\begin{tabular}{lccr}
\toprule
period & first paper & last paper \\
\midrule
1 &  1805.08929 &  1810.00786 \\
2 &  1810.00787 &  1902.00889 \\ 
3 &  1902.00890 &  1905.13537 \\
4 &  1905.13538 &  1909.11935 \\
5 &  1909.11936 &  2001.06560 \\
6 &  2001.06561 &  2005.02178 \\
7 &  2005.02179 &  2008.04251 \\
8 &  2008.04252 &  2011.09225 \\
9 &  2011.09226 &  2103.01828 \\
10 &  2103.01829 &  2106.04209 \\
11 &  2106.04210 &  2109.09152 \\
12 &  2109.09153 &  2112.12197 \\
13 &  2112.12198 &  2204.01835 \\
14 &  2204.01836 &  2207.06075 \\
15 &  2207.06076 &  2210.10618 \\
16 &  2210.10619 &  2301.10909 \\
17 &  2301.10910 &  2304.13927 \\
18 &  2304.13928 &  2307.10978 \\
19 &  2307.10979 &  2310.09716 \\
20 &  2310.09717 &  2401.02417 \\
\bottomrule
\end{tabular}
\end{center}
\end{table}

\section{arXiv categories}
Formally, arXiv has 8 categories in total: physics, mathematics, computer science, quantitative biology, quantitative finance, statistics, electrical engineering and systems science, economics. The first 3 categories contribute the vast majority of arXiv articles, around 91\% among the 1 million articles. Hence, we divided the physics papers into sub-categories: astrophysics, condensed matter, high energy physics, etc. The four categories (computer science, mathematics, astrophysics, condensed matter) we selected account for 70\% of the total number of articles. To avoid repetition, we also only count the first category of the article for those that have multiple categories (cross-postings).

\section{Other observations}
We define the change factor in the frequency of word $i$, $R_i$, as follows: 
\begin{equation}
    R_i = \frac{\max_{t}(f_i(t)) - \min_{t}(f_i(t))}{\max_{t}(f_i(t))}
\end{equation}
where $f_i(t)$ is the count of word $i$ during the time period $t$. 

Similarly, we define a change factor in the frequency of word $i$, ${R_i}'$:
\begin{equation}
    {R_i}' = \frac{\max_{t}({f_i}'(t)) - \min_{t}({f_i}'(t))}{\max_{t}({f_i}'(t))}
\end{equation}
where ${f_i}'(t)$ is the count of word $i$ in period $t$, normalized to the same value of $\sum_if_i(t)$ for all periods $t$.

\begin{figure}[h]
    \centering
    \begin{subfigure}[b]{0.48\textwidth}
        \centering
        \includegraphics[width=\textwidth]{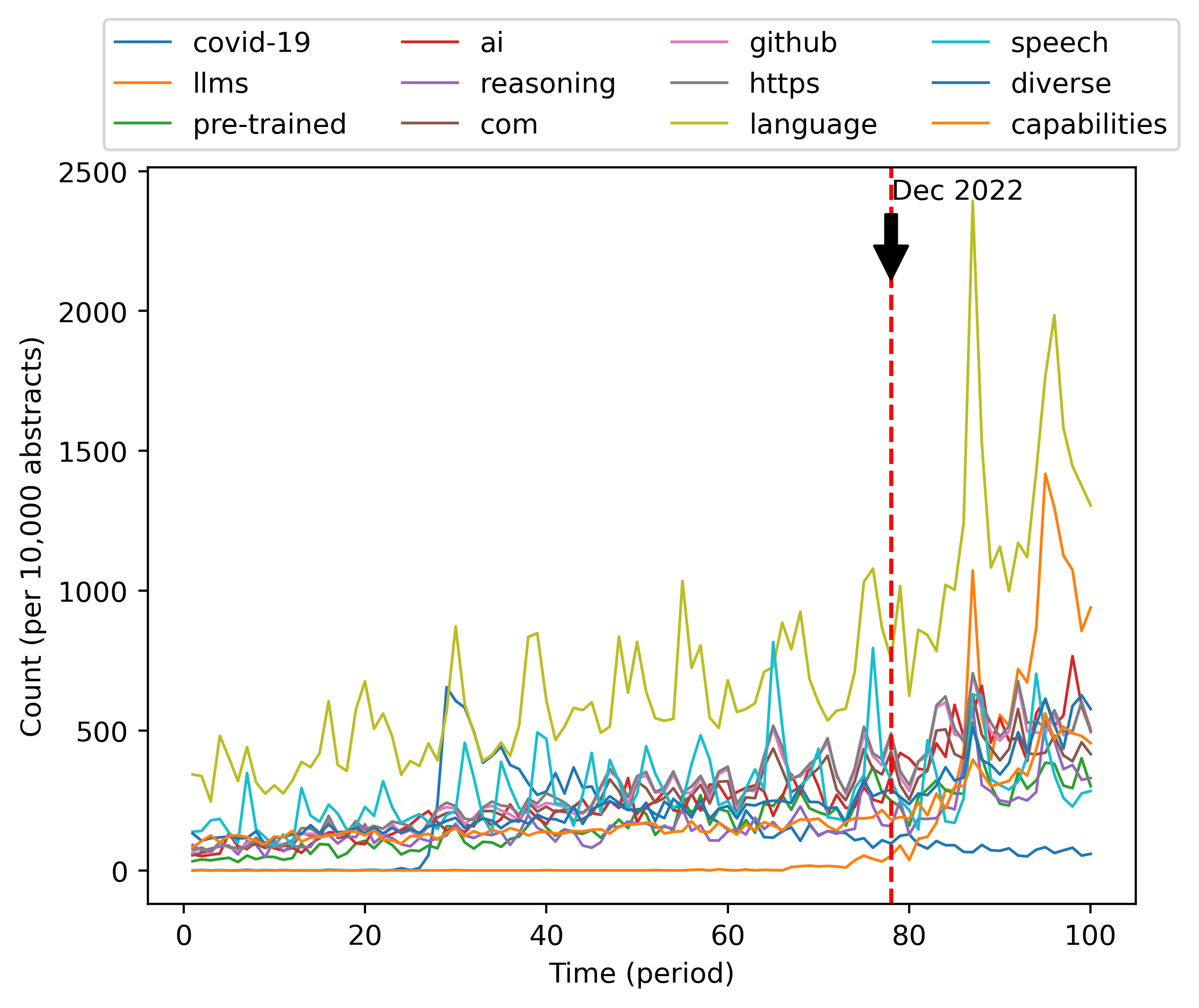}
        \caption{The 12 words with the highest change rate $R_i$ and satisfying  $\max_{t}(f_i(t))>500$.}
        \label{ex}
    \end{subfigure}
        \hfill
    \begin{subfigure}[b]{0.48\textwidth}
        \centering
        \includegraphics[width=\textwidth]{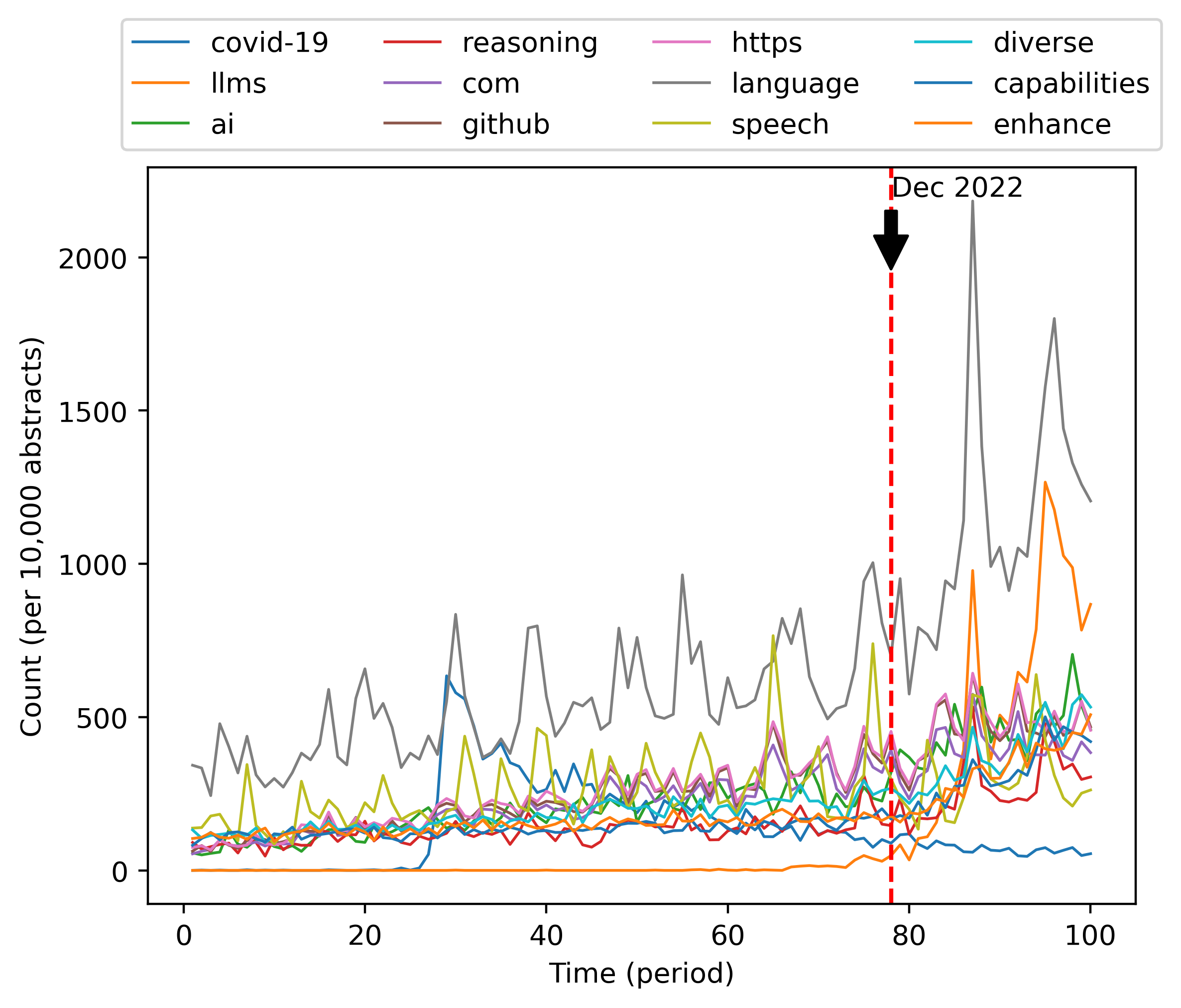}
        \caption{The 12 words with the highest change rate ${R_i}'$ and satisfying  $\max_{t}({f_i}'(t))>500$.}
        \label{obs_1}
    \end{subfigure}
    \caption{Words with the highest change rate in frequency}
\end{figure}

Figure \ref{ex} and Figure \ref{obs_1} illustrate that most of the words with the largest change rate in the time period considered (generally, an increase) in the abstracts are related to hot research topics of the last few years, such as ``Covid-19'', ``LLMs'', ``AI''.

\begin{figure}[h]
    \centering
    \begin{subfigure}[b]{0.48\textwidth}
        \centering
        \includegraphics[width=\textwidth]{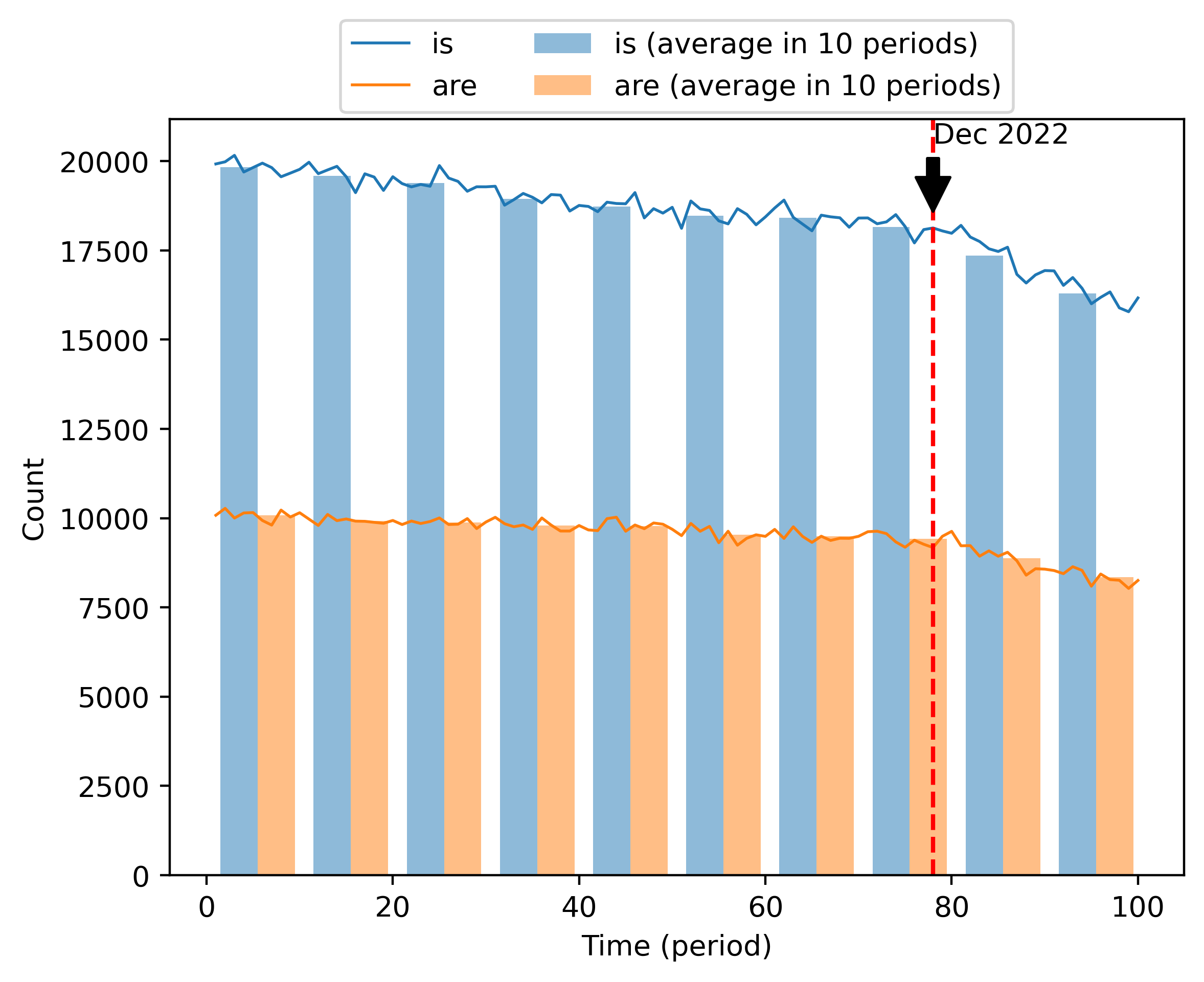}
    \end{subfigure}
        \hfill
    \begin{subfigure}[b]{0.48\textwidth}
        \centering
        \includegraphics[width=\textwidth]{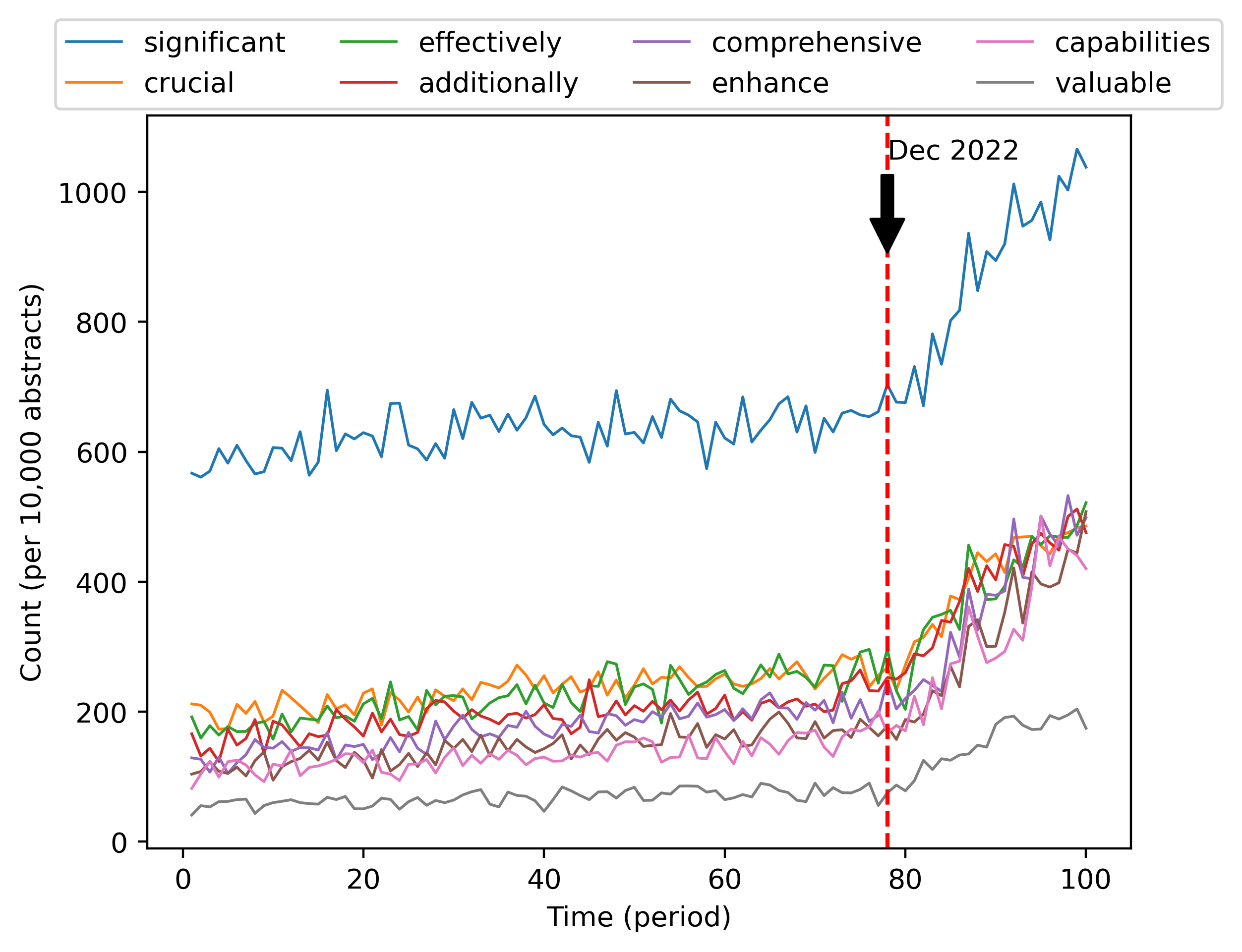}
    \end{subfigure}
    \caption{Word frequency changes (with different normalization) in abstracts.} \label{norm_ex}
\end{figure}

The total number of words in all abstracts of the first period is used as a base to normalize the frequency of words in the other periods, and the corresponding results are shown Figure \ref{norm_ex}.

\section{Correlation between simulated and real data}
We also defined the word frequency change in all abstracts from year $t-1$ to year $t$, $R_{ij,t}$:
\begin{equation}
    R_{ij,t} = \frac{F_{ij,t}-F_{ij,t-1}}{F_{ij,t-1}} \, ,
\end{equation}
where $F_{ij,t}$ represent frequency of word $i$ per arXiv abstract in category $j$ in year $t$. 

\begin{figure}[h]
    \centering
    \begin{subfigure}[b]{0.48\textwidth}
        \centering
        \includegraphics[width=\textwidth]{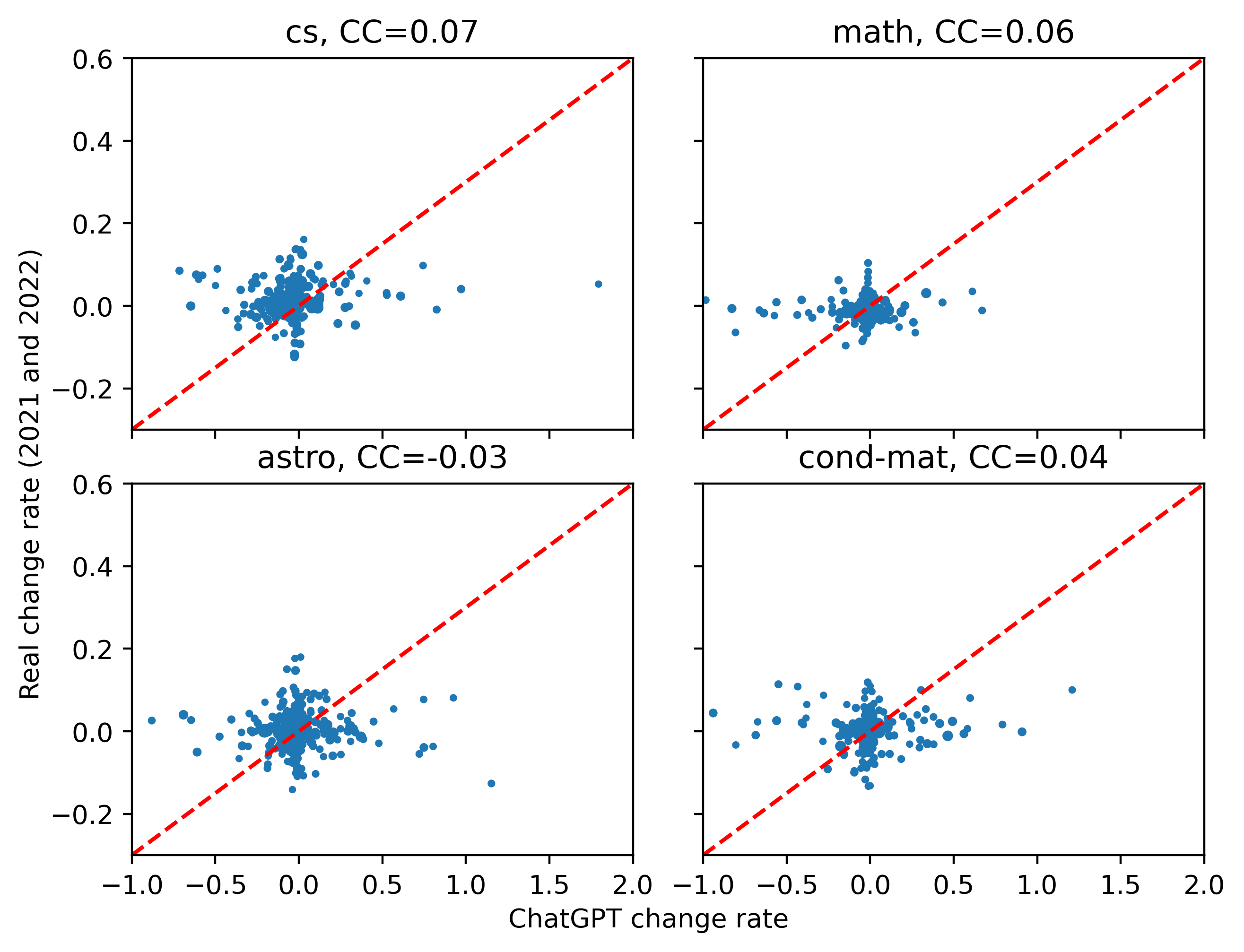}
        \caption{From 2021 to 2022.}
        \label{change_compare_2122}
    \end{subfigure}
    \hfill
    \begin{subfigure}[b]{0.48\textwidth}
        \centering
        \includegraphics[width=\textwidth]{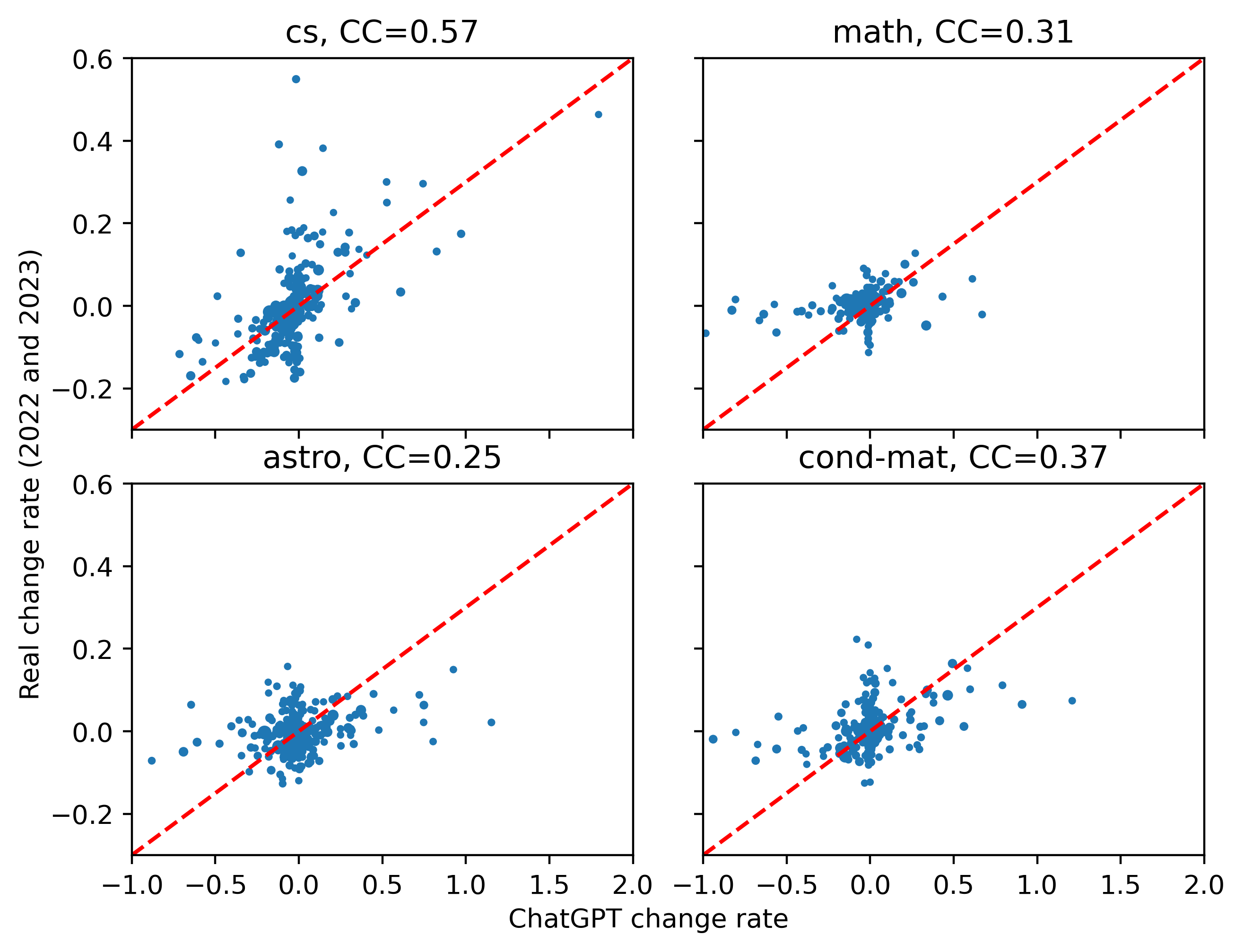}
        \caption{From 2022 to 2023.}
        \label{change_compare}
    \end{subfigure}
    \caption{Comparison of the predicted frequency change rate due to ChatGPT $\hat{r}_{ij}$ (x-axis) and the actual word frequency change for all abstracts (y-axis). CC indicates the correlation coefficient.}
    \label{wc_change_com}
\end{figure}

Only words with a frequency larger than 0.1 times per abstract before ChatGPT processing are plotted in Figure \ref{change_compare_2122} and Figure \ref{change_compare}. The correlation coefficient between the word frequency change in arXiv abstracts and our estimated ChatGPT-induced word frequency change is very small in all four categories of abstracts, as shown in 
Figure \ref{change_compare_2122}.

However, Figure \ref{change_compare} presents a totally different pattern, where $\hat{r}_{ij}$ and $R_{ij,2023}$ are strongly correlated, especially in computer science abstracts. Although many words seem insensitive to ChatGPT, we can still see a positive correlation for some words in this figure, even among the other categories.

\section{Parameters}
\subsection{ChatGPT simulations}
\begin{itemize}
    \item model: gpt-3.5-turbo-1106
    \item temperature: 0.7
    \item seed: 1106
    \item top\_p: 0.2
\end{itemize}
\subsection{Calibration}
\begin{itemize}
    \item $\displaystyle \frac{1}{q^d_{ij}}$: 10, 20, 30, 40, 50, 60, 70, 80, 100, 150, 200, 500
    \item $\hat{r}_{ij}$: 0.1, 0.15, 0.2, 0,3, 0.4, 0.5, 0.6, 0.7, 0.8 (corresponding value of $\displaystyle \frac{\hat{r}_{ij}+1}{\hat{r}^2_{ij}}$)
\end{itemize}

For example, when we take $\displaystyle \frac{1}{q^d_{ij}}<10$ and $\displaystyle \frac{\hat{r}_{ij}+1}{\hat{r}^2_{ij}} <\frac{0.1+1}{0.1^2}$ for abstracts in computer science, the words that satisfy the conditions are: 'the', 'is', 'for', 'by', 'be', 'this', 'are', 'i', 'at', 'which', 'an', 'have', 'but', 'we', 'all', 'they', 'one', 'has', 'their', 'other', 'there', 'more', 'new', 'any', 'these', 'time', 'than', 'some', 'only', 'two', 'into', 'them', 'our', 'under', 'first', 'most', 'then', 'over', 'work', 'where', 'many', 'through', 'well', 'how', 'even', 'while', 'however', 'high', 'given', 'present', 'large', 'research', 'different', 'set', 'study', 'important', 'several', 'e', 'further', 'including', 'often', 'provide', 'due', 'using', 'better', 'various', 'problem', 'show', 'problems', 'design', 'proposed', 'g', 'across', 'approach', 'existing', 'compared', 'task', 'learn', 'improve', 'achieve', 'novel', 'domain', 'demonstrate', 'introduce', 'propose', 'prediction'.

And when $\displaystyle \frac{1}{q^d_{ij}}<50$ and $\displaystyle \frac{\hat{r}_{ij}+1}{\hat{r}^2_{ij}} <\frac{0.8+1}{0.8^2}$, the words are: 'i', 'would', 'so', 'some', 'what', 'out', 'work', 'very', 'because', 'much', 'good', 'way', 'great', 'here', 'since', 'might', 'last', 'end', 'means', 'having', 'thus', 'above', 'give', 'e', 'further', 'far', 'find', 'although', 'show', 'n', 'help', 'together', 'particular', 'whose', 'issue', 'according', 'addition', 'usually', 'art', 'especially', 'respect', 'works', 'shows', 'g', 'makes', 'hard', 'significant', 'run', 'address', 'particularly', 'idea', 'consider', 'includes', 'built', 'adopted', 'obtain', 'establish', 'useful', 'leading', 'performed', 'create', 'named', 'conducted', 'resulting', 'hence', 'findings', 'towards', 'prove', 'build', 'perform', 'moreover', 'describe', 'besides', 'demonstrated', 'via', 'presents', 'mainly', 'fail', 'namely', 'allowing', 'demonstrate', 'advances', 'suffer', 'overcome', 'introduce', 'accurately', 'identifying', 'enhance', 'crucial', 'etc', 'utilize', 'demonstrates', 'additionally', 'focuses', 'motivated', 'characterize'. 

\section{Noise analysis}
\subsection{Variance in real data}
\begin{figure}[h]
    \centering
    \begin{subfigure}[b]{0.48\textwidth}
        \centering
        \includegraphics[width=\textwidth]{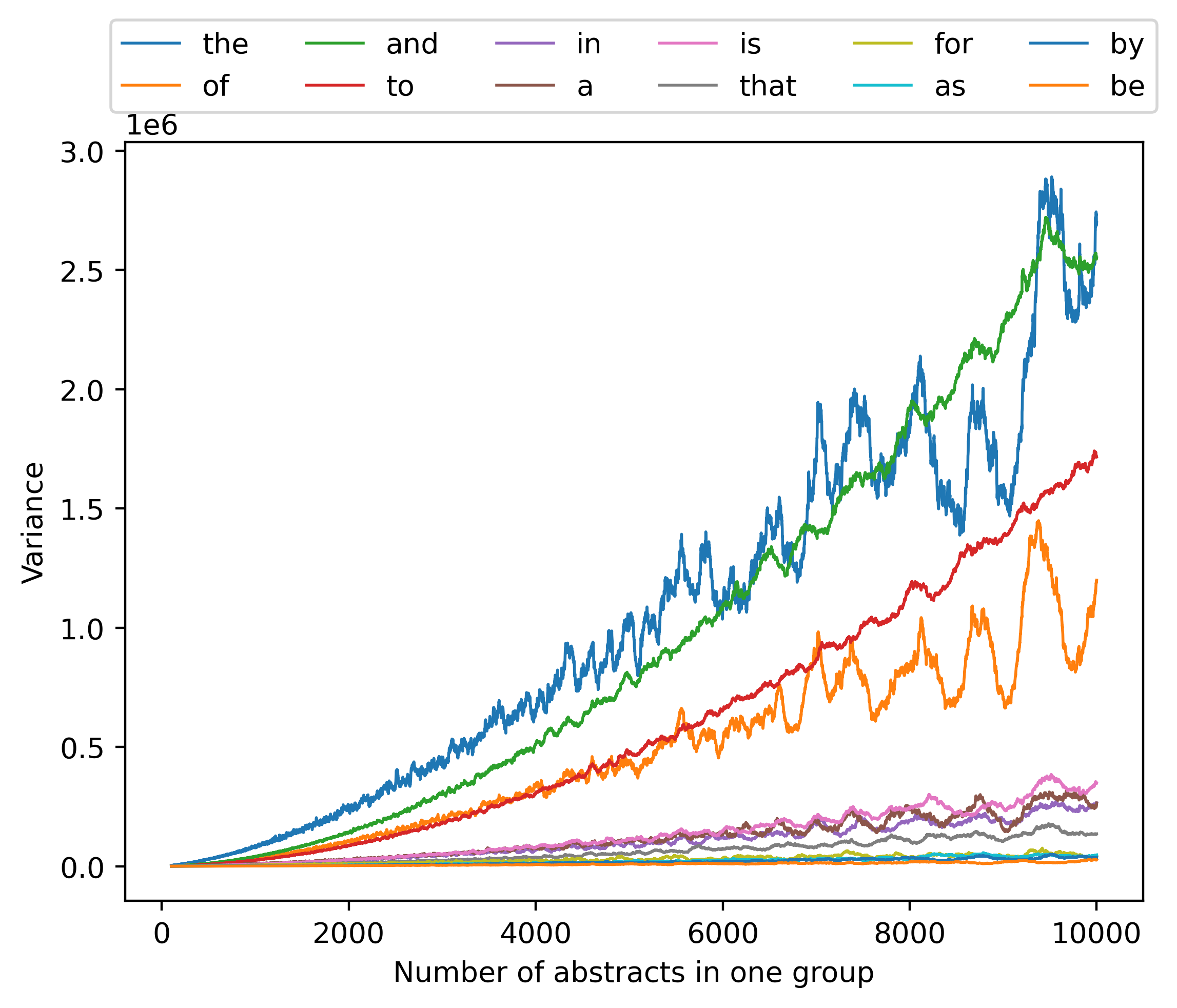}
        \caption{Variance of the word counts.}
        \label{var_cs}
    \end{subfigure}
    \hfill
    \begin{subfigure}[b]{0.48\textwidth}
        \centering
        \includegraphics[width=\textwidth]{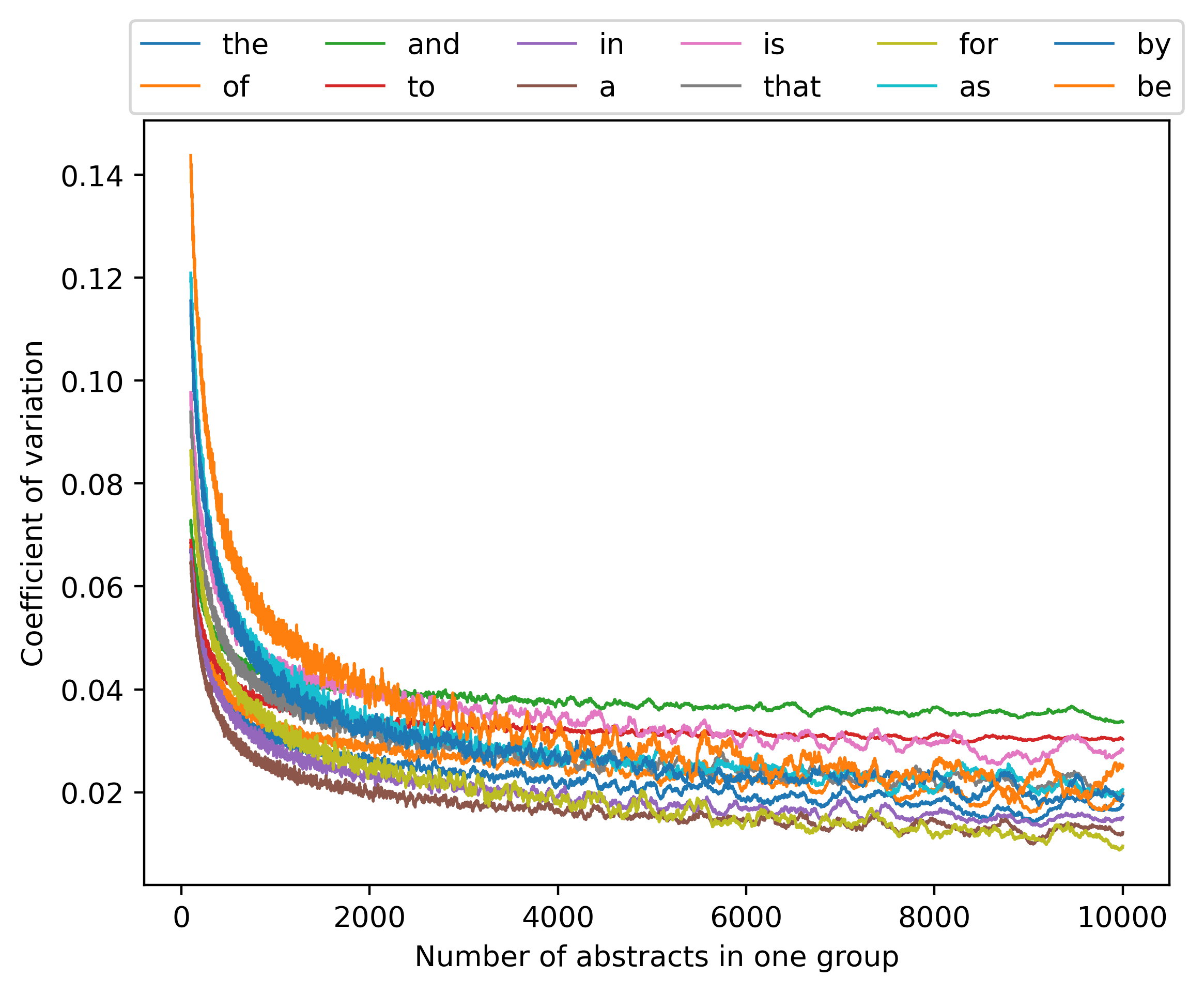}
        \caption{Coefficient of variation.}
        \label{coeff_var_cs}
    \end{subfigure}
    \caption{Variance of the 12 most frequent words.}
\end{figure}

Abstracts in the \textit{cs} category among the first 500,000 articles were divided into groups in chronological order, with the same number in each group. We counted the number of occurrences of each word within each group, and calculated the variance between the different groups. This was repeated as a function of the number of abstracts included in each group, and the results are shown in Figure \ref{var_cs}.

\begin{figure}[h]
    \centering
    \begin{subfigure}[b]{0.48\textwidth}
        \centering
        \includegraphics[width=\textwidth]{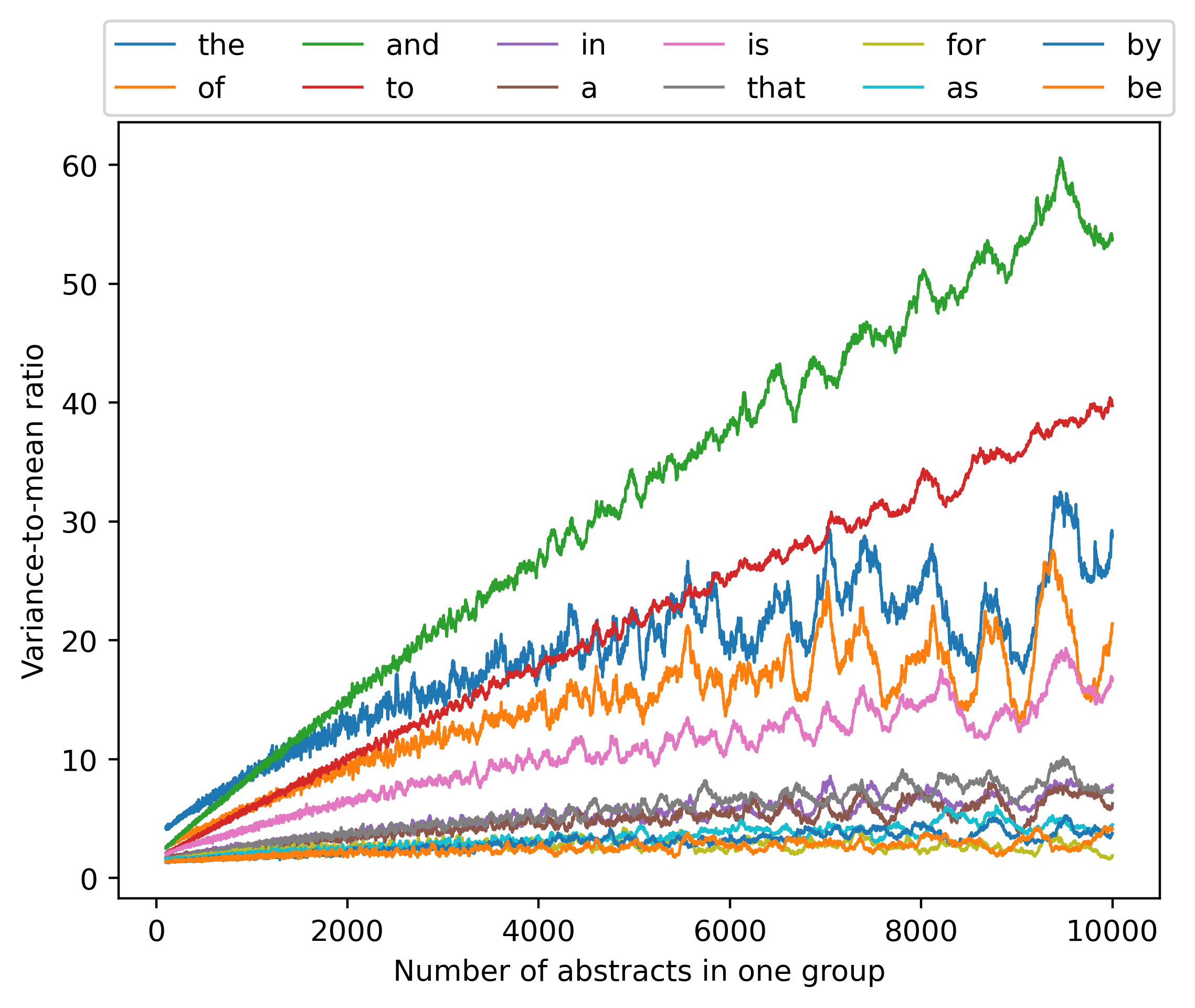}
    \end{subfigure}
        \hfill
    \begin{subfigure}[b]{0.48\textwidth}
        \centering
        \includegraphics[width=\textwidth]{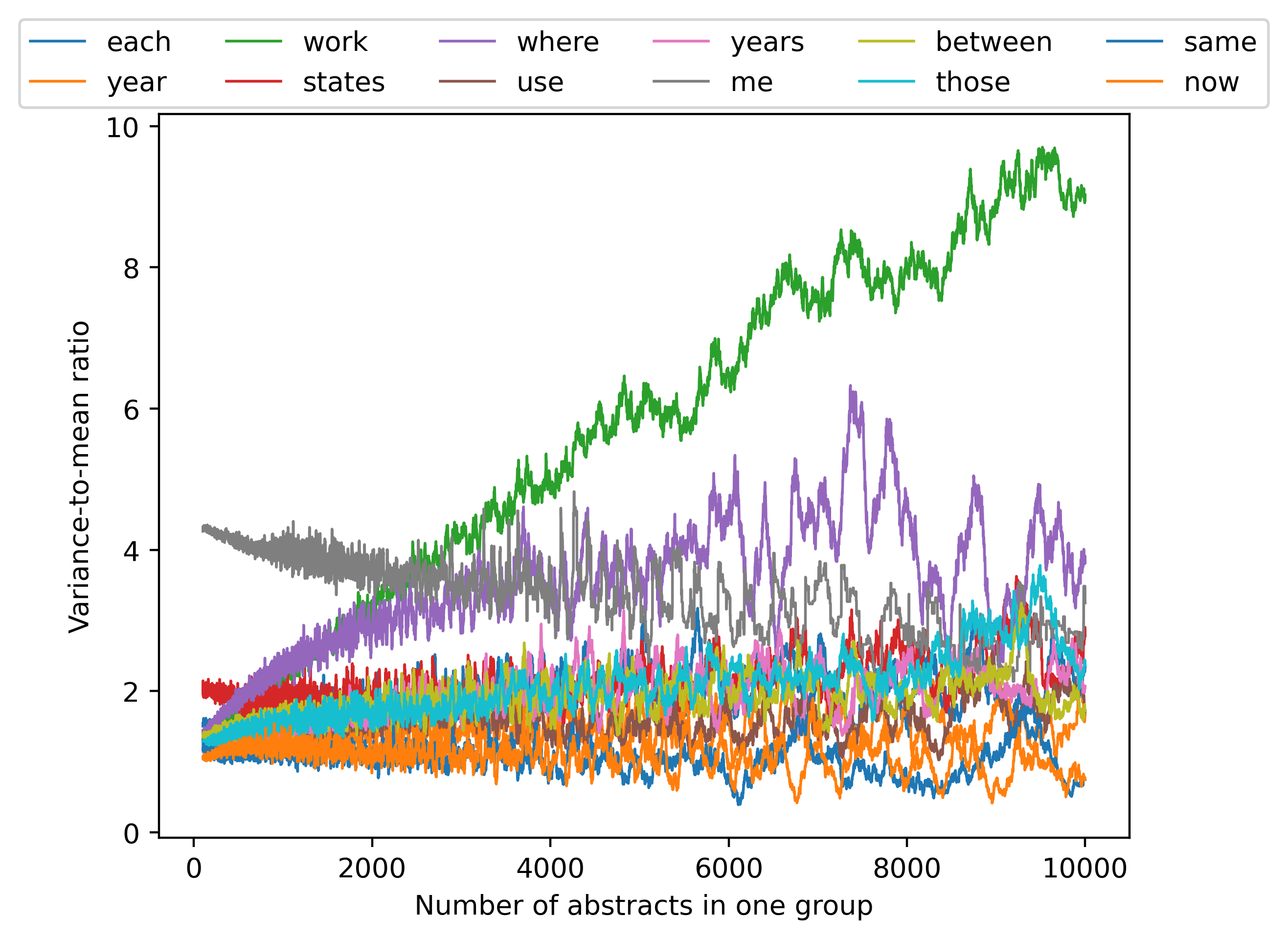}
    \end{subfigure}
    \caption{Variance-to-mean ratio}\label{var_mean}
\end{figure}

Then we also analyzed the coefficient of variation (defined as the standard deviation of the sum divided by the mean of the sum) for the 12 most frequent words, as shown in Figure \ref{coeff_var_cs}, and the variance-to-mean ratio (defined as the variance of the sum of a word's counts divided by the mean of the sum), as shown in Figure \ref{var_mean}.

We observe that, at least for a subset of the words considered here, the variance-to-mean ratios are essentially on the same scale (although there are words that do not follow this pattern). Therefore, a simple Gaussian distribution
\begin{equation}
    \delta_{ij}(f_{ij}) \sim \mathcal {N}(0 ,f_{ij} \sigma_{ij}^{2}) \, .
\end{equation}
which corresponds to case 2, seems to be a reasonable approximation. 

\subsection{Calculation details}
\paragraph{Case 1:}
Therefore, all terms on the right-hand side of Eq. (\ref{bias_eta_g}) are zero-mean noise, except for the last one:
\begin{equation}
g_{ij}(t) \frac{\partial g_{ij}(t)}{\partial\eta_j(t)}
        =  g_{ij}(t) \frac{f^*_{ij}(t)(\eta_j(t)f^*_{ij}(t)+2\delta_{ij}(*))\epsilon_{ij}(1)}{2(\eta_j(t)f^*_{ij}(t)+ \delta_{ij}(*))^{\frac{3}{2}}}- g_{ij}(t)\frac{f^*_{ij}\epsilon^s_{ij}(1)}{\sqrt{q^d_{ij}}} \, .
\end{equation}
Removing the items with zero means, we get
\begin{equation}
         \mathrm{E}\left[g_{ij}(t) \frac{\partial g_{ij}(t)}{\partial\eta_j(t)}\right] =  \frac{\eta_j(t)(f^*_{ij}(t))^2(\eta_j(t)f^*_{ij}(t)+2\delta_{ij}(*))\sigma^2_{ij,\epsilon}}{2(\eta_j(t)f^*_{ij}(t)+ \delta_{ij}(*))^{2}}  + \frac{\eta_j(t) (f^*_{ij}(t))^2\sigma^2_{ij,\epsilon}}{q^d_{ij}} \, .
\label{eq:eta_bias_e_expression}
\end{equation}
\paragraph{Case 2:}
We can define $g^c_{ij}(t)$ and $\xi^c_{ij}(t)$:
\begin{align}
\begin{split}
    g^c_{ij}(t) = & \eta_j(t)f^*_{ij}(t)\epsilon^{\eta}_{ij}(q,f,t)+ \sqrt{\eta_j(t)f^*_{ij}(t)}(\hat{r}_{ij}+\epsilon^{\eta}_{ij}(q,f,t))\delta_{ij}(1)
\end{split}
  \\
  \xi^c_{ij}(t) = & \sqrt{f^*_{ij}(t)}\delta'_{ij}(1)
\end{align}
As $\xi^c_{ij}(t)$ doesn't depend on $\eta_j(t)$, the loss function under this assumption is:
\begin{equation}
        L^c_{j,t,g}(\eta_j)
        = \frac{1}{n_j}\sum_{i \in I_j} ( h_{ij}(t) - \eta_{j}(t)x_{ij}(t)-g^c_{ij}(t))^2 =\frac{1}{n_j}\sum_{i \in I_j}(\xi^c_{ij}(t))^2 \, . 
\end{equation}
And we will get a complex expression for the bias part like Eq. (\ref{bias_eta_g}).

As in case 1, we set $\frac{\partial L^c_{j,t,g}(\eta_j)}{\partial \eta_j} =0$ to obtain the new estimate $\hat{\eta}^g_j(t)$ corrected for bias and noise,
\begin{equation}
    \begin{split}
        (\hat{\eta}^g_j(t) - \hat{\eta}_j(t))\sum_{i \in I_j} x^2_{ij}(t)
        = &\sum_{i \in I_j} \frac{\partial g^c_{ij}(t)}{\partial\eta_j(t)} \left( h_{ij}(t) - \eta_{j}(t)x_{ij}(t)\right)  \\ 
        & - \sum_{i \in I_j} x_{ij}(t)g^c_{ij}(t) -\sum_{i \in I_j}g^c_{ij}(t)\frac{\partial g^c_{ij}(t)}{\partial\eta_j(t)} 
    \end{split}
\end{equation}
where
\begin{equation}
    \begin{split}
        \frac{\partial g^c_{ij}(t)}{\partial\eta_j(t)} = & f^*_{ij}(t)\epsilon^{\eta}_{ij}(q,f,t) + \eta_j(t)f^*_{ij} \frac{\partial \epsilon^{\eta}_{ij}(q,f,t)}{\partial \eta_j(t)} + \frac{\sqrt{f^*_{ij}(t)}}{2\sqrt{\eta_j(t)}}(\hat{r}_{ij}+\epsilon^{\eta}_{ij}(q,f,t))\delta_{ij}(1) \\ 
    & + \sqrt{\eta_j(t)f^*_{ij}(t)}\frac{\partial \epsilon^{\eta}_{ij}(q,f,t)}{\partial \eta_j(t)}\delta_{ij}(1) \, .
    \end{split} 
\end{equation}
The bias part is also expressed as
\begin{equation} 
\hat{\eta}_j(t) - \hat{\eta}^g_j(t) = 
    \frac{\sum_{i \in I_j}\mathrm{E}\left[g^c_{ij}(t) \frac{\partial g^c_{ij}(t)}{\partial\eta_j(t)}\right]}{\sum_{i \in I_j} (f^*_{ij} (t)\hat{r}_{ij})^2} \, .
\end{equation}
Also with the same assumptions for $\epsilon_{ij}(\cdot)$ and $\epsilon^s_{ij}(\cdot)$, $\epsilon_{ij}(f_{ij}) \sim \mathcal {N}(0 ,f_{ij} \sigma_{ij,\epsilon}^{2})$ and $\epsilon^s_{ij}(f_{ij}) \sim \mathcal {N}(0 ,f_{ij} \sigma_{ij,\epsilon}^{2})$. then we can obtain an expression for $\epsilon^{\eta}_{ij}(q,f,t)$,
\begin{equation}
    \epsilon^{\eta}_{ij}(q,f,t) =  \frac{\epsilon_{ij}(1)}{\sqrt{\eta_j(t)f^*_{ij}(t)+ \sqrt{\eta_j(t)f^*_{ij}(t)}\delta_{ij}(1)}}- \frac{\epsilon^s_{ij}(1)}{\sqrt{q^d_{ij}}} 
\end{equation}
and its derivative,
\begin{equation}
        \frac{\partial \epsilon^{\eta}_{ij}(q,f,t) }{\partial \eta_j(t)} =  \frac{-\left(2f^*_{ij}(t)\sqrt{\eta_j(t)}+\sqrt{f^*_{ij}(t)}\delta_{ij}(1)\right)\epsilon_{ij}(1)}{4\sqrt{\eta_j(t)}\left(\eta_j(t)f^*_{ij}(t)+ \sqrt{\eta_j(t)f^*_{ij}(t)}\delta_{ij}(1)\right)^{\frac{3}{2}}} \, .
\end{equation}
Combining the above equations, we can get similar conclusions as in case 1.

\end{document}